\documentclass{article} 
\usepackage{iclr2017_conference,times}
\usepackage{hyperref}
\usepackage{url}
\usepackage{graphicx} 
\usepackage{subfigmat}
\usepackage{amsfonts}
\usepackage{amssymb}
\usepackage{amsmath}
\usepackage{etoolbox} 
\graphicspath{ {./figures/}}

\title{Revisiting Small Batch Training for \\Deep Neural Networks}

\author{Dominic Masters and Carlo Luschi \\
Graphcore Research\\
Bristol, UK \\
\texttt{\{dominicm,carlo\}@graphcore.ai} \\
}

\begin{document}
\setlength{\abovedisplayskip}{12pt plus 2pt minus 2pt}
\setlength{\belowdisplayskip}{12pt plus 2pt minus 2pt}
\setlength{\abovedisplayshortskip}{8pt plus 2pt minus 2pt}
\setlength{\belowdisplayshortskip}{8pt plus 2pt minus 2pt}
\addtolength{\jot}{6pt}
\maketitle

\begin{abstract}
Modern deep neural network training is typically based on mini-batch stochastic gradient optimization. While the use of large mini-batches increases the available computational parallelism, small batch training has been shown to provide improved generalization performance and allows a significantly smaller memory footprint, which might also be exploited to improve machine throughput.

In this paper, we review common assumptions on learning rate scaling and training duration, as a basis for an experimental comparison of test performance for different mini-batch sizes.
We adopt a learning rate that corresponds to a constant average weight update per gradient calculation (i.e., per unit cost of computation), and point out that this results in a variance of the weight updates that increases linearly with the mini-batch size $m$.

The collected experimental results for the CIFAR-10, CIFAR-100 and ImageNet datasets show that increasing the mini-batch size progressively reduces the range of learning rates that provide stable convergence and acceptable test performance. On the other hand, small mini-batch sizes provide more up-to-date gradient calculations, which yields more stable and reliable training. The best performance has been consistently obtained for mini-batch sizes between $m = 2$ and $m = 32$, which contrasts with recent work advocating the use of mini-batch sizes in the thousands.
\end{abstract}

\section{Introduction}

The use of deep neural networks has recently enabled significant advances in a number of applications, including computer vision, speech recognition and natural language processing, and in reinforcement learning for robotic control and game playing~\citep{LeCun15,Schmidhuber15,Goodfellow16,Arulkumaran17}.

Deep learning optimization is typically based on Stochastic Gradient Descent (SGD) or one of its variants~\citep{Bottou16,Goodfellow16}.
The SGD update rule relies on a stochastic approximation of the expected value of the gradient of the loss function over the training set, based on a small subset of training examples, or {\em mini-batch}.

The recent drive to employ progressively larger batch sizes is motivated by the desire to improve the parallelism of SGD, both to increase the efficiency of current processors and to allow distributed processing on a larger number of nodes~\citep{Dean12,Das16}.
In contrast, the use of small batch sizes has been shown to improve generalization performance and optimization convergence~\citep{Wilson03,LeCun12,Keskar16}. The use of small batch sizes also has  the advantage of requiring a significantly smaller memory footprint, which affords an opportunity to design processors which gain efficiency by exploiting memory locality. This motivates a fundamental trade-off in the choice of batch size.

\citet{Hoffer17} have shown empirically that it is possible to maintain generalization performance with large batch training by performing the same number of SGD updates. However, this implies a computational overhead proportional to the mini-batch size, which negates the effect of improved hardware efficiency due to increased parallelism.

In order to improve large batch training performance for a given training computation cost, it has been proposed to scale the learning rate linearly with the batch size~\citep{Krizhevsky14,Chen16,Bottou16,Goyal17}.
In this work, we suggest that the discussion about the scaling of the learning rate with the batch size depends on how the problem is formalized, and revert to the view of~\citet{Wilson03}, that considers the SGD weight update formulation based on the sum, instead of the average, of the gradients over the mini-batch. From this perspective, using a fixed learning rate keeps the expectation of the weight update per training example constant for any choice of batch size.
At the same time, as will be discussed in Section~\ref{sec:Background}, holding the expected value of the weight update per gradient calculation constant while increasing the batch size implies a linear increase of the variance of the weight update with the batch size.

To investigate these issues, we have performed a comprehensive set of experiments for a range of network architectures.
The results provide evidence that increasing the batch size results in both a degradation of the test performance and a progressively smaller range of learning rates that allows stable training, where the notion of stability here refers to the robust convergence of the training algorithm.

The paper is organized as follows. Section~\ref{sec:Background} briefly reviews the main work on batch training and normalization. 
Section~\ref{sec:Batch_Train_Performance} presents a range of experimental results on training and generalization performance for the CIFAR-10, CIFAR-100 and ImageNet datasets.
Previous work has compared training performance using batch sizes of the order of 128--256 with that of very large batch sizes of up to 4096~\citep{Hoffer17} or 8192~\citep{Goyal17}, while we consider the possibility of using a wider range of values, down to batch sizes as small as 2 or 4. 
Section~\ref{sec:Discussion} provides a discussion of the main results presented in the paper.
Finally, conclusions are drawn in Section~\ref{sec:Conclusions}.

\section{Background: Batch Training and Batch Normalization}
\label{sec:Background}

\subsection{Stochastic Gradient Optimization}
\label{sec:SGD}

We assume a deep network model with parameters $\boldsymbol{\theta}$, and consider the non-convex optimization problem corresponding to the minimization of the loss function $L(\boldsymbol{\theta})$, with respect to $\boldsymbol{\theta}$.  $L(\boldsymbol{\theta})$ is defined as the sample average of the loss per training example $L_i(\boldsymbol{\theta})$ over the training set,
\begin{equation}\label{eq:Loss}
L(\boldsymbol{\theta}) = \frac{1}{M} \sum_{i=1}^{M} L_i(\boldsymbol{\theta}) \, ,
\end{equation}
where $M$ denotes the size of the training set. The above empirical loss is used as a proxy for the expected value of the loss with respect to the true data generating distribution.

Batch gradient optimization originally referred to the case where the gradient computations were accumulated over one presentation of the entire training set before being applied to the parameters, and stochastic gradient methods were typically online methods with parameter update based on a single training example. Current deep learning stochastic gradient algorithms instead use parameter updates based on gradient averages over small subsets of the full training set, or {\em mini-batches}, and the term {\em batch size} is commonly used to refer to the size of a mini-batch~\citep{Goodfellow16}.

Optimization based on the SGD algorithm uses a stochastic approximation of the gradient of the loss $L(\boldsymbol{\theta})$ obtained from a mini-batch $\mathcal{B}$ of $m$ training examples, resulting in the weight update rule
\begin{gather}
\boldsymbol{\theta}_{k+1} = \boldsymbol{\theta}_k + \eta \;  \Delta \boldsymbol{\theta}_k  \label{eq:SGD_1}\\
\Delta \boldsymbol{\theta}_k = - \frac{1}{m} \sum_{i=1}^{m} \nabla_{\boldsymbol{\theta}} L_i(\boldsymbol{\theta}_k)  \label{eq:SGD_2}
\end{gather}
where $\eta$ denotes the learning rate.

From (\ref{eq:SGD_1}), (\ref{eq:SGD_2}) the mean value of the SGD weight update is given by $\mathbb{E}\{ \eta \, \Delta \boldsymbol{\theta} \} = -\eta \, \mathbb{E}\{ \nabla_{\boldsymbol{\theta}} L(\boldsymbol{\theta}) \}$. Therefore, for a given batch size $m$, the expected value of the weight update per unit cost of computation (per training example, i.e., per gradient calculation $\nabla_{\boldsymbol{\theta}} L_i(\boldsymbol{\theta})$) is proportional to $\eta/m$:
\begin{equation} \label{eq:LR_1}
\frac{1}{m} \, \mathbb{E}\{ \eta \, \Delta \boldsymbol{\theta} \} = - \frac{\eta}{m} \, \mathbb{E}\{ \nabla_{\boldsymbol{\theta}} L(\boldsymbol{\theta}) \} \, .
\end{equation}
This implies that, when increasing the batch size, a linear increase of the learning rate $\eta$ with the batch size $m$ is required to keep the mean SGD weight update per training example constant.

This {\em linear scaling} rule has been widely adopted, e.g., in~\citet{Krizhevsky14}, \citet{Chen16}, \citet{Bottou16}, \citet{Smith17} and \citet{Jastrzebski17}.

On the other hand, as shown in~\citet{Hoffer17}, when $m \ll M$, the covariance matrix of the weight update $\mathrm{Cov}\{ \eta \, \Delta \boldsymbol{\theta} \}$ scales linearly with the quantity $\eta^2/m \, $.

This implies that, adopting the linear scaling rule, an increase in the batch size would also result in a linear increase in the covariance matrix of the weight update $\eta \, \Delta \boldsymbol{\theta}$.
Conversely, to keep the scaling of the covariance of the weight update vector $\eta \, \Delta \boldsymbol{\theta}$ constant would require scaling $\eta$ with the square root of the batch size $m$~\citep{Krizhevsky14,Hoffer17}.

\subsection{A Different Perspective on Learning Rate Scaling}
\label{sec:Perspective}

We suggest that the discussion about the linear or sub-linear increase of the learning rate with the batch size is the purely formal result of assuming the use of the \textit{average} of the local gradients over a mini-batch in the SGD weight update.

As discussed in~\citet{Wilson03}, current batch training implementations typically use a weight correction based on the average of the local gradients according to equation (\ref{eq:SGD_2}), while earlier work on batch optimization often assumed a weight correction based on the {\em sum} of the local gradients~\citep{Wilson03}. Using the sum of the gradients at the point $\boldsymbol{\theta}_k$, the SGD parameter update rule can be rewritten as
\begin{equation} \label{eq:SGD_3}
\boldsymbol{\theta}_{k+1} = \boldsymbol{\theta}_k - \tilde{\eta} \; \sum_{i=1}^{m} \nabla_{\boldsymbol{\theta}} L_i(\boldsymbol{\theta}_k)
\end{equation}
and in this case, if the batch size $m$ is increased, the mean SGD weight update per training example is kept constant by simply maintaining a constant learning rate $\tilde{\eta}$. This is equivalent to using the linear scaling rule.

In the SGD weight update formulation (\ref{eq:SGD_3}), the learning rate $\tilde{\eta}$ corresponds to the {\em base learning rate} that would be obtained from a linear increase of the learning rate $\eta$ of (\ref{eq:SGD_1}), (\ref{eq:SGD_2}) with the batch size $m$, i.e.
\begin{equation} \label{eq:LR_3}
\tilde{\eta} = \eta \, \frac{1}{m} \, .
\end{equation}

At the same time, from Section~\ref{sec:SGD} the variance of the weight update scales linearly with $\tilde{\eta}^2 \! \cdot m$. Therefore keeping the base learning rate $\tilde{\eta}$ constant implies a linear increase of the variance with the batch size $m$.

If we now consider a sequence of updates from the point $\boldsymbol{\theta}_k$ with a batch size $m$, from (\ref{eq:SGD_3}) the value of the weights at step $k+n$ is expressed as
\begin{equation} \label{eq:SGD_4}
\boldsymbol{\theta}_{k+n} = \boldsymbol{\theta}_k - \tilde{\eta} \; \sum_{j=0}^{n-1}  \sum_{i=1}^{m} \nabla_{\boldsymbol{\theta}} L_{i+jm}(\boldsymbol{\theta}_{k+j}) \, ,
\end{equation}
while increasing the batch size by a factor $n$ implies that the weights at step $k+1 \,$ (corresponding to the same number of gradient calculations) are instead given by
\begin{equation} \label{eq:SGD_5}
\boldsymbol{\theta}_{k+1} = \boldsymbol{\theta}_k - \tilde{\eta} \;  \sum_{i=1}^{n m} \nabla_{\boldsymbol{\theta}} L_i(\boldsymbol{\theta}_k) \, .
\end{equation}
The comparison of (\ref{eq:SGD_4}), (\ref{eq:SGD_5}) highlights how, under the assumption of constant $\tilde{\eta}$, large batch training can be considered to be an approximation of small batch methods that trades increased parallelism for stale gradients.

From (\ref{eq:SGD_4}), (\ref{eq:SGD_5}), the end point values of the weights $\boldsymbol{\theta}_{k+n}$ and $\boldsymbol{\theta}_{k+1}$ after $n m$ gradient calculations will generally be different, except in the case where one could assume 
\begin{equation} \label{eq:SGD_6}
\nabla_{\boldsymbol{\theta}} L_i(\boldsymbol{\theta}_k) \approx \nabla_{\boldsymbol{\theta}} L_i(\boldsymbol{\theta}_{k+j}) \, , \qquad j = 1, ..., n - 1
\end{equation}
as also pointed out in~\citet{Wilson03} and~\citet{Goyal17}.
Therefore, for larger batch sizes, the update rule will be governed by progressively different dynamics for larger values of $n$, especially during the initial part of the training, during which the network parameters are typically changing rapidly over the non-convex loss surface~\citep{Goyal17}.
It is important to note that (\ref{eq:SGD_6}) is a better approximation if the batch size and/or the base learning rate are small.


The purely formal difference of using the average of the local gradients instead of the sum has favoured the conclusion that using a larger batch size could provide more `accurate' gradient estimates and allow the use of larger learning rates. 
However, the above analysis shows that, from the perspective of maintaining the expected value of the weight update per unit cost of computation, this may not be true. In fact, using smaller batch sizes allows gradients based on more up-to-date weights to be calculated, which in turn allows the use of higher base learning rates, as each SGD update has lower variance. Both of these factors potentially allow for faster and more robust convergence.

\subsection{Effect of Batch Normalization}
\label{sec:Batch_Norm}

The training of modern deep networks commonly employs {\em Batch Normalization}~\citep{Ioffe15}. This technique has been shown to significantly improve training performance and has now become a standard component of many state-of-the-art networks.

Batch Normalization (BN) addresses the problem of {\em internal covariate shift} by reducing the dependency of the distribution of the input activations of each layer on all the preceding layers.
This is achieved by normalizing activation $x_i$ for each feature,
\begin{equation}
\hat{x}_i = \frac{\, x_i-\hat{\mu}_\mathcal{B}}{\sqrt{\, \hat{\sigma}_{\mathcal{B}}^2 + \epsilon \,}}, 
\end{equation}
where $\hat{\mu}_\mathcal{B}$ and $\hat{\sigma}_{\mathcal{B}}^2$ denote respectively the sample mean and sample variance calculated over the batch $\mathcal{B}$ for one feature.
The normalized values $\hat{x}_i$ are then further scaled and shifted by the learned parameters  $\gamma$ and $\beta$~\citep{Ioffe15}
\begin{equation}
y_i = \hat{x}_i \cdot \gamma + \beta \, .
\end{equation}

For the case of a convolutional layer, with a feature map of size $p \times q$ and batch size $m$, the sample size for the estimate of $\hat{\mu}_\mathcal{B}$ and $\hat{\sigma}_{\mathcal{B}}^2$ is given by $m \cdot p \cdot q$, while for a fully-connected layer the sample size is simply equal to $m$. For very small batches, the estimation of the batch mean and variance can be very noisy, which may limit the effectiveness of BN in reducing the covariate shift.
Moreover, as pointed out in~\citet{Ioffe17}, with very small batch sizes the estimates of the batch mean and variance used during training become a less accurate approximation of the mean and variance used for testing. The influence of BN on the performance for different batch sizes is investigated in Section \ref{sec:Batch_Train_Performance}.

We observe that the calculation of the mean and variance across the batch makes the loss calculated for a particular example dependent on other examples of the same batch. This intrinsically ties the empirical loss (\ref{eq:Loss}) that is minimized to the choice of batch size. In this situation, the analysis of (\ref{eq:SGD_4}), (\ref{eq:SGD_5}) in Section~\ref{sec:Perspective} is only strictly applicable for a fixed value of the batch size used for BN. 
In some cases the overall SGD batch is split into smaller sub-batches used for BN. A common example is in the case of distributed processing, where BN is often implemented independently on each separate worker to reduce communication costs (e.g. \citet{Goyal17}). For this scenario, the discussion relating to (\ref{eq:SGD_4}), (\ref{eq:SGD_5}) in Section~\ref{sec:Perspective} is directly applicable assuming a fixed batch size for BN. The effect of using different batch sizes for BN and for the weight update of the optimization algorithm will be explored in Section~\ref{sec:Dist_Performance}.

Different types of normalization have also been proposed~\citep{Ba16,Salimans16,Arpit16,Ioffe17,Wu18}. In particular, {\em Batch Renormalization}~\citep{Ioffe17} and {\em Group Normalization}~\citep{Wu18} have reported improved performance for small batch sizes.

\subsection{Other Related Work}
\label{sec:Other_Work}

\citet{Hoffer17} have shown empirically that the reduced generalization performance of large batch training is connected to the reduced number of parameter updates over the same number of epochs (which corresponds to the same computation cost in number of gradient calculations).
\citet{Hoffer17} present evidence that it is possible to achieve the same generalization performance with large batch size, by increasing the training duration to perform the same number of SGD updates. Since from (\ref{eq:SGD_2}) or (\ref{eq:SGD_3}) the number of gradient calculations per parameter update is proportional to the batch size $m$, this implies an increase in computation proportional to $m$.

In order to reduce the computation cost for large batches due to the required longer training, \citet{Goyal17} have investigated the possibility of limiting the training length to the same number of epochs. To this end, they have used larger learning rates with larger batch sizes based on the linear scaling rule, which from (\ref{eq:SGD_3}) is equivalent to keeping constant the value of $\tilde{\eta}$. As discussed in Section~\ref{sec:Perspective}, this implies an increase of the variance of the weight update linearly with the batch size. 
The large batch training results of~\citet{Goyal17} report a reduction in generalization performance using the linear scaling rule with batch sizes up to $8192$.
However, \citet{Goyal17} have found that the test performance of the small batch ($m = 256$) baseline could be recovered using a {\em gradual warm-up} strategy. We explore the impact of this strategy in Section~\ref{sec:Performance_BN_WU} and find that it does not fully mitigate the degradation of the generalization performance with increased batch size.

\citet{Jastrzebski17} claim that both the SGD generalization performance and training dynamics are controlled by a noise factor given by the ratio between the learning rate and the batch size, which corresponds to linearly scaling the learning rate. The paper also suggests that the invariance to the simultaneous rescaling of both the learning rate and the batch size breaks if the learning rate becomes too large or the batch size becomes too small. However, the evidence presented in this  paper only shows that the linear scaling rule breaks when it is applied to progressively larger batch sizes.

\citet{Smith17} have recently suggested using the linear scaling rule to increase the batch size instead of decreasing the learning rate during training. While this strategy simply results from a direct application of (\ref{eq:LR_1}) and (\ref{eq:LR_3}), and guarantees a constant mean value of the weight update per gradient calculation, as discussed in Section~\ref{sec:Perspective} decreasing the learning rate and increasing the batch size are only equivalent if one can assume (\ref{eq:SGD_6}), which does not generally hold for large batch sizes. It may however be approximately applicable in the last part of the convergence trajectory, after having reached the region corresponding to the final minimum. Finally, it is worth noting that in practice increasing the batch size during training is more difficult than decreasing the learning rate, since it may require, for example, modifying the computation graph.

\section{Batch Training Performance}
\label{sec:Batch_Train_Performance}

This section presents numerical results for the training performance of convolutional neural network architectures for a range of batch sizes $m$ and base learning rates $\tilde{\eta}$.

The experiments have been performed on three different datasets: CIFAR-10, CIFAR-100~\citep{Krizhevsky09} and ImageNet~\citep{Krizhevsky12}. The CIFAR-10 and CIFAR-100 experiments have been run for different AlexNet and ResNet models~\citep{Krizhevsky12,He15} exploring the main factors that affect generalization performance, including BN, data augmentation and network depth. Further experiments have investigated the training performance of the ResNet-50 model~\citep{He15} on the ImageNet dataset, to confirm significant conclusions on a more complex task.

\subsection{Implementation Details}
\label{sec:Implementation}

For all the experiments, training has been based on the standard SGD optimization. While SGD with momentum  \citep{Sutskever13} has often been used to train the network models considered in this work \citep{Krizhevsky12, He15, Goyal17}, it has been shown that the optimum momentum coefficient is dependent on the batch size \citep{Smith17a,Goyal17}. For this reason momentum was not used, with the purpose of isolating the interaction between batch size and learning rate.

The numerical results for different batch sizes have been obtained by running experiments for a range of base learning rates $\tilde{\eta} = \eta / m$, from $2^{-12}$ to $2^{0}$. 
For all learning rates and all batch sizes, training has been performed over the same number of epochs, which corresponds to a constant computational complexity (a constant number of gradient calculations).

For the experiments using the CIFAR-10 and CIFAR-100 datasets, we have investigated a reduced version of the AlexNet model of~\citet{Krizhevsky12}, and the ResNet-8, ResNet-20 and ResNet-32 models as described in \citet{He15}. The reduced AlexNet implementation uses convolutional layers with stride equal to 1, kernel sizes equal to $[5,3,3,3,3]$, number of channels per layer equal to $[16,48,96,64,64]$, max pool layers with $3\times3$ kernels and stride 2, and 256 hidden nodes per fully-connected layer. Unless stated otherwise, all the experiments have been run for a total of 82 epochs for CIFAR-10 and 164 epochs for CIFAR-100, with a learning rate schedule based on a reduction by a factor of 10 at 50\% and at 75\% of the total number of iterations. Weight decay has been also applied, with $\lambda = 5 \cdot 10^{-4}$ for the AlexNet model~\citep{Krizhevsky12} and $\lambda = 10^{-4}$ for the ResNet models~\citep{He15}. In all cases, the weights have been initialized using the method described in \citet{He15a}.

The CIFAR datasets are split into a 50,000 example training set and a 10,000 example test set~\citep{Krizhevsky09}, from which we have obtained the reported results. The experiments have been run with and without augmentation of the training data. This has been performed following the procedure of~\citet{Lin13} and~\citet{He15}, which pads the images by 4 pixels on each side, takes a random 32$\times$32 crop, and then randomly applies a horizontal flip.

The ImageNet experiments were based on the ResNet-50 model of~\citet{He15}, with preprocessing that follows the method of \citet{Simonyan14} and \citet{He15}. As with the CIFAR experiments, the weights have been initialized based on the method of~\citet{He15a}, with a weight decay parameter $\lambda = 10^{-4}$. Training has been run for a total of 90 epochs, with a learning rate reduction by a factor of 10 at 30, 60 and 80 epochs, following the procedure of \citet{Goyal17}. 

For all the results, the reported test or validation accuracy is the median of the final 5 epochs of training~\citep{Goyal17}.

\subsection{Performance Without Batch Normalization}
\label{sec:Performance_noBN}

\begin{figure}[p]
\centering
\patchcmd{\subfigmatrix}{\hfill}{\hspace{0.6cm}}{}{} 
\begin{subfigmatrix}{2}
\subfigure{\includegraphics[width=0.45\linewidth]{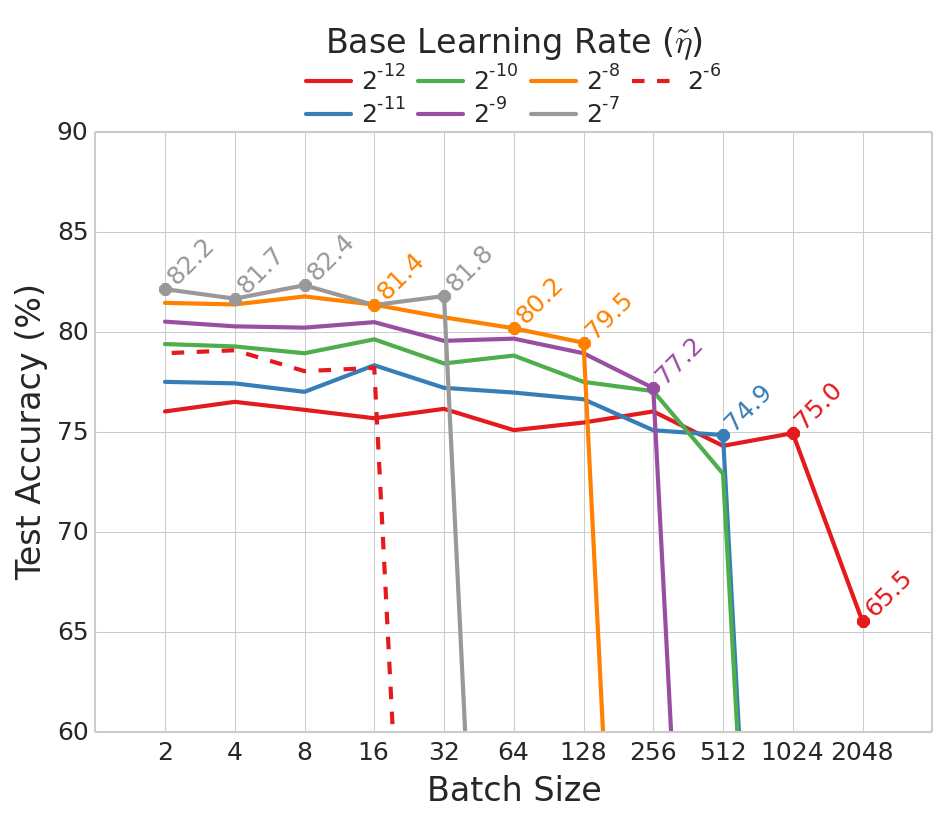}}
\subfigure{\includegraphics[width=0.45\linewidth]{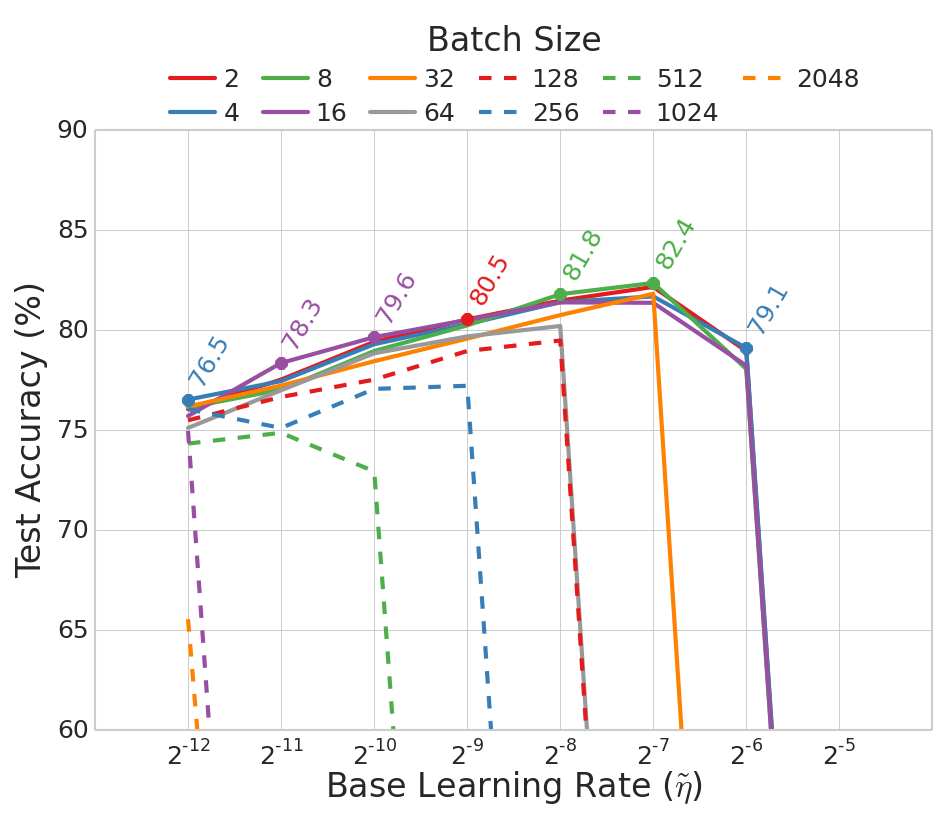}}
\end{subfigmatrix}
\caption{Test performance of reduced AlexNet model without BN, for different batch sizes and increasing values of $\tilde{\eta} = \eta/m$. CIFAR-10 dataset without data augmentation.}
\label{fig:AlexNet_CIFAR-10_noBN}

\vspace{0.5cm}

\begin{subfigmatrix}{2}
\subfigure{\includegraphics[width=0.45\linewidth]{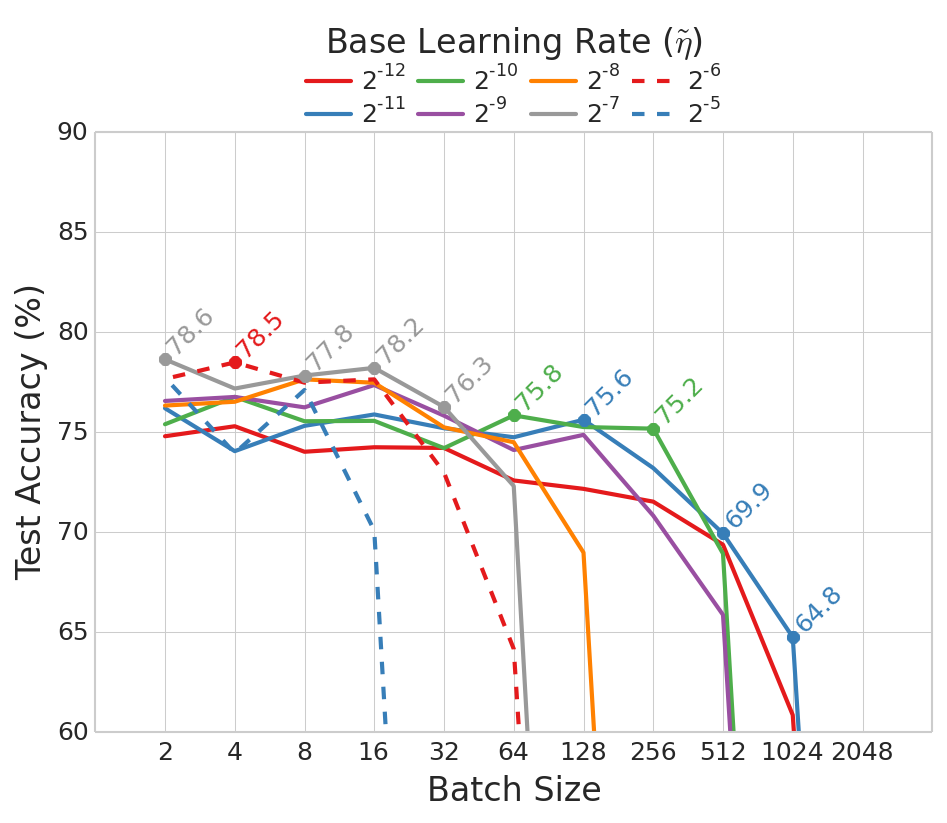}}
\subfigure{\includegraphics[width=0.45\linewidth]{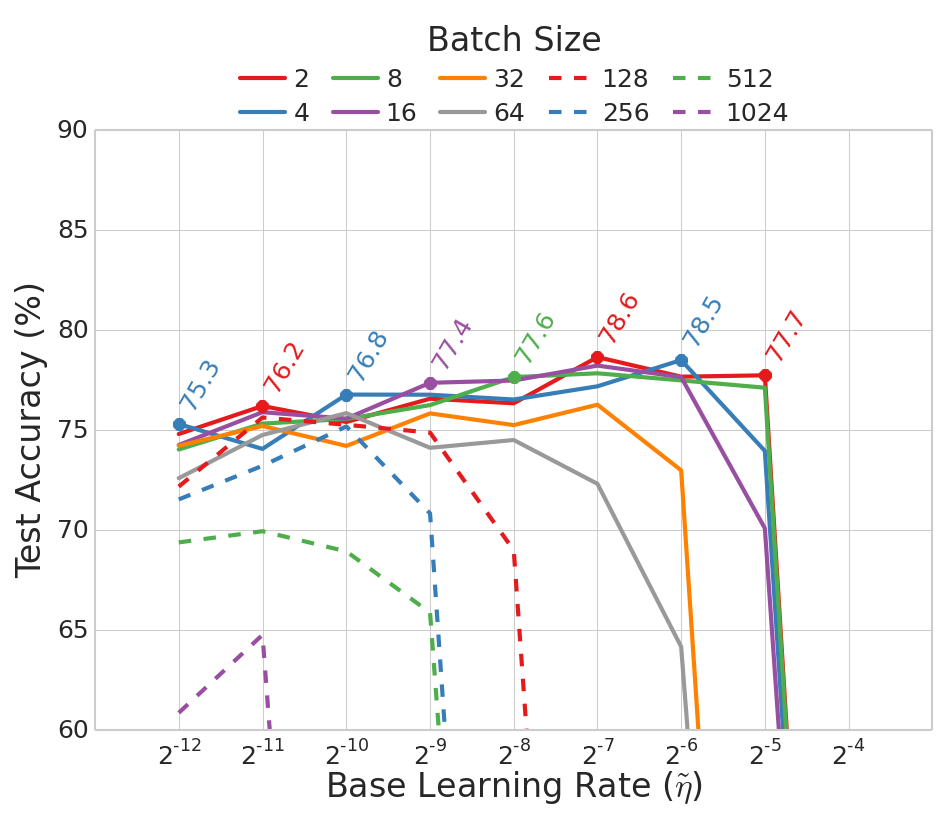}}
\end{subfigmatrix}
\caption{Test performance of ResNet-8 model without BN, for different batch sizes and increasing values of $\tilde{\eta} = \eta/m$. CIFAR-10 dataset without data augmentation.}
\label{fig:ResNet-8_CIFAR-10_noAug_noBN}

\vspace{0.5cm}

\begin{subfigmatrix}{2}
\subfigure{\includegraphics[width=0.45\linewidth]{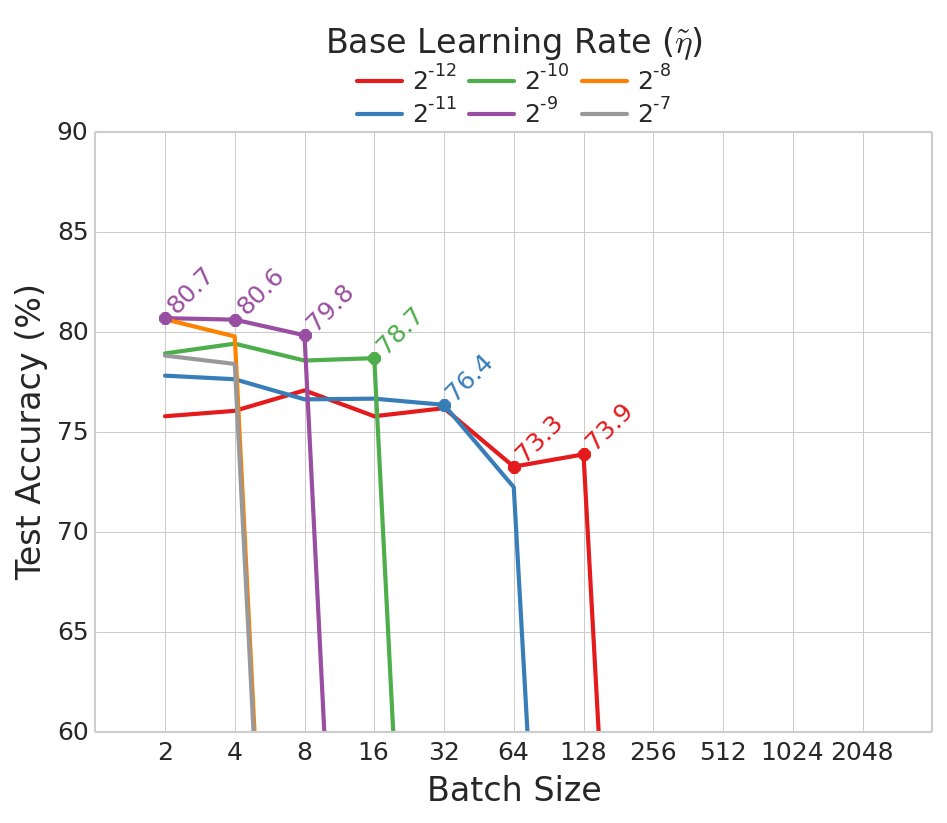}}
\subfigure{\includegraphics[width=0.45\linewidth]{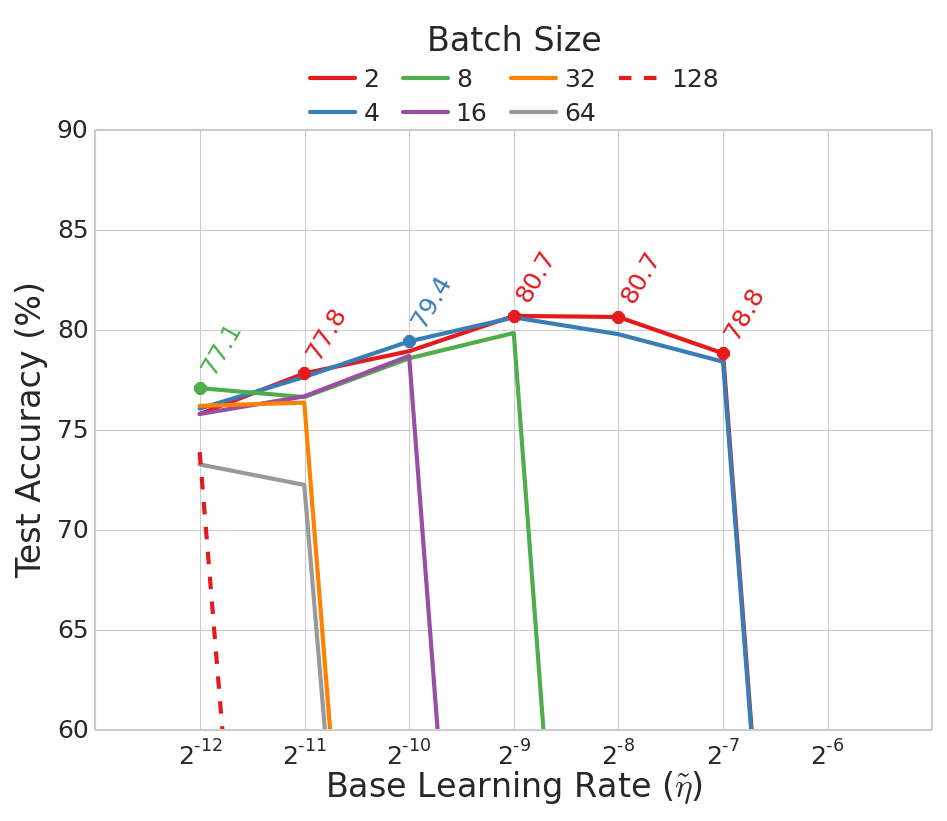}}
\end{subfigmatrix}
\caption{Test performance of ResNet-20 model without BN, for different batch sizes and increasing values of $\tilde{\eta} = \eta/m$. CIFAR-10 dataset without data augmentation.}
\label{fig:ResNet-20_CIFAR-10_noAug_noBN}
\end{figure}

Figures~\ref{fig:AlexNet_CIFAR-10_noBN}--\ref{fig:ResNet-20_CIFAR-10_noAug_noBN} report the CIFAR-10 test performance of the reduced AlexNet, ResNet-8 and ResNet-20 models without BN or data augmentation, for  different values of the batch size $m$ and base learning rate $\tilde{\eta} = \eta / m$. 
The results show a clear performance improvement for progressively smaller batch sizes: for the reduced AlexNet model, the best test accuracy is obtained for batch sizes $m = 8$ or smaller, while for the ResNet models the best accuracy is obtained for batch sizes $m = 4$ and $m = 2$. This agrees with the expectation of Section~\ref{sec:Perspective}.

The figures also indicate the presence of an optimum value of the base learning rate $\tilde{\eta} = \eta / m$ that delivers stable training only for batch sizes smaller than a critical value. Increasing the batch size with a constant base learning rate $\tilde{\eta}$ (equivalent to a linear scaling of $\eta$ with the batch size $m$) also results in a reduction in generalization performance followed by a sharp decrease as training becomes unstable.

On the other hand, both the convergence robustness and the corresponding generalization performance consistently improve by reducing the batch size.
From Figures~\ref{fig:AlexNet_CIFAR-10_noBN}--\ref{fig:ResNet-20_CIFAR-10_noAug_noBN} we observe that the use of small batch sizes corresponds to the largest range of learning rates that provides stable convergence.
Comparing ResNet-8 and ResNet-20 it appears that that using smaller batch sizes reduces possible training difficulties associated with the use of deeper models.
Not shown in the figures is the case of ResNet-32 without BN, for which all the CIFAR-10 experiments with base learning rate $\tilde{\eta}$ from $2^{-12}$ to $2^{0}$ and batch size from $m = 16$ to $m = 2048$ failed to converge, and learning was successful only for the cases of batch size $m = 2$,  $m = 4$ and $m = 8$, with a best test performance of 76\%.


\subsection{Performance With Batch Normalization}
\label{sec:Performance_BN}

We now present experimental results with BN.
As discussed in Section~\ref{sec:Batch_Norm}, BN has been shown to significantly improve the training convergence and performance of deep networks.

\begin{figure}[tb]
\centering
\includegraphics[width=0.50\linewidth]{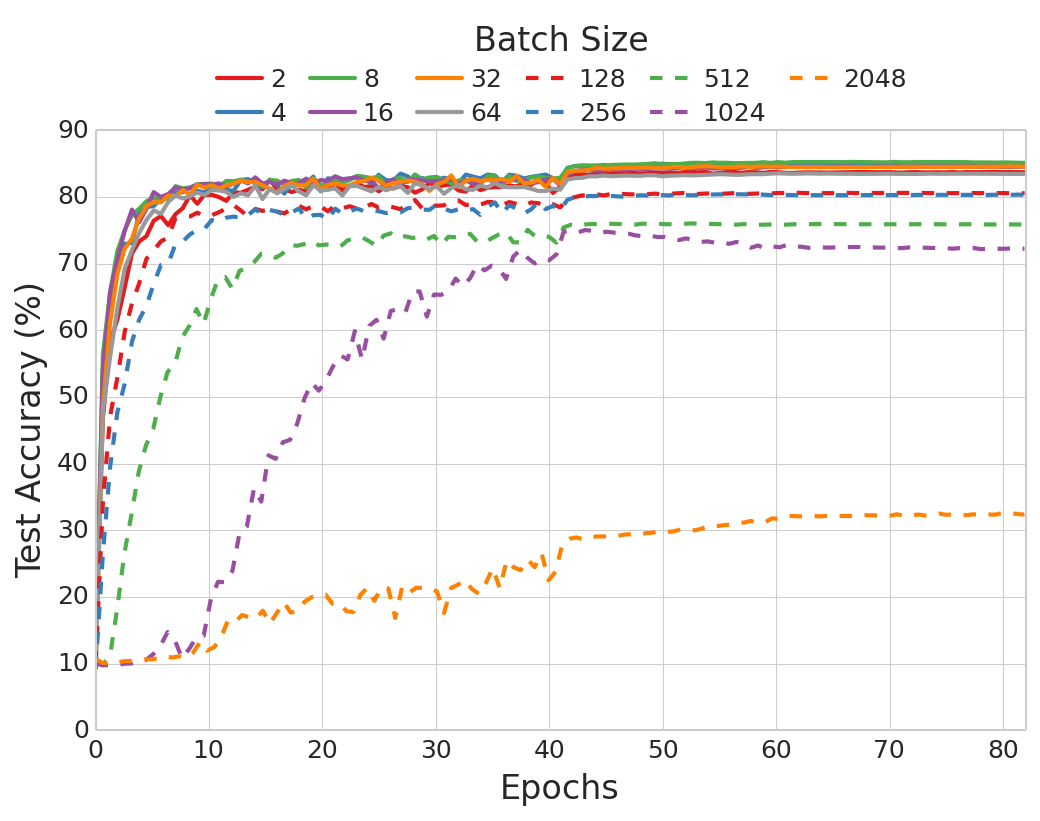}
\caption{Convergence curves for ResNet-32 model with BN, for $\tilde{\eta} = 2^{-8}$ and for different values of the batch size. CIFAR-10 dataset without data augmentation.}
\label{fig:ResNet32_CIFAR-10_training_BN}
\end{figure}

Figure~\ref{fig:ResNet32_CIFAR-10_training_BN} shows the convergence curves for CIFAR-10, ResNet-32 training with BN, for different batch sizes. The curves have been obtained by keeping the base learning rate $\tilde{\eta}$ constant (which is equivalent to applying a linear scaling of $\eta$ with the batch size $m$). From the plots, we observe similar convergence for some of the curves corresponding to the same value of $\tilde{\eta}$ with batch sizes in the range $4 \leq m \leq 64$, but with degraded training performance for larger batch sizes. Not shown here, analogous plots for higher learning rates show similar training divergence for large batch sizes.

\begin{figure}[p]
\centering
\patchcmd{\subfigmatrix}{\hfill}{\hspace{0.6cm}}{}{} 
\begin{subfigmatrix}{2}
\subfigure{\includegraphics[width=0.45\linewidth]{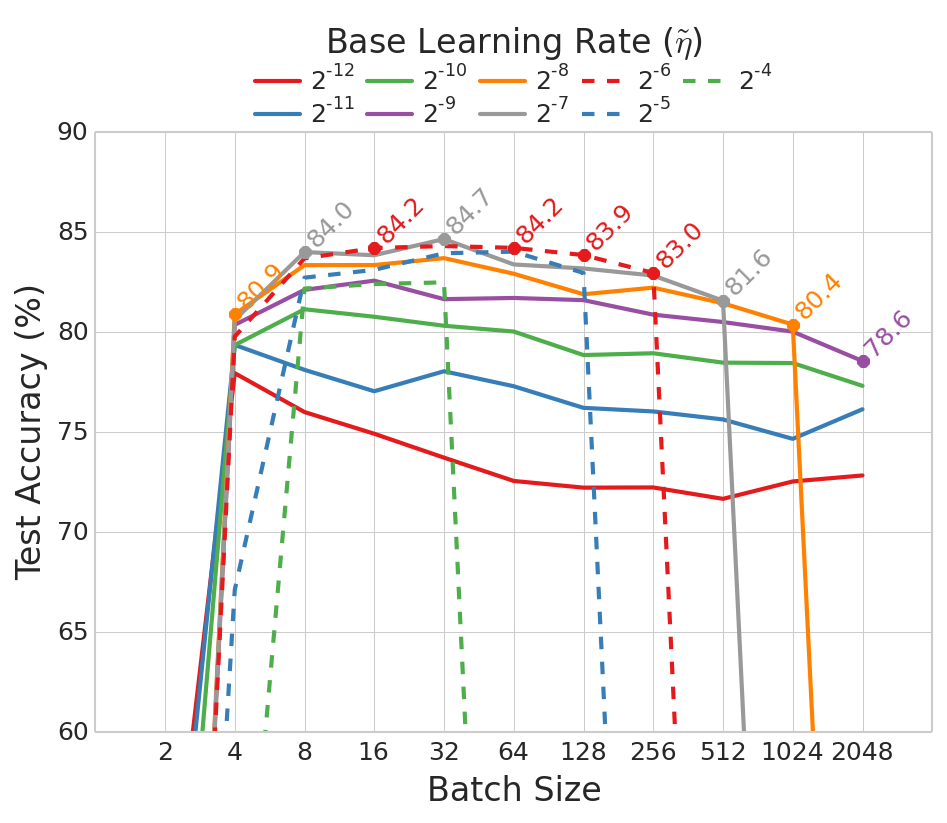}}
\subfigure{\includegraphics[width=0.45\linewidth]{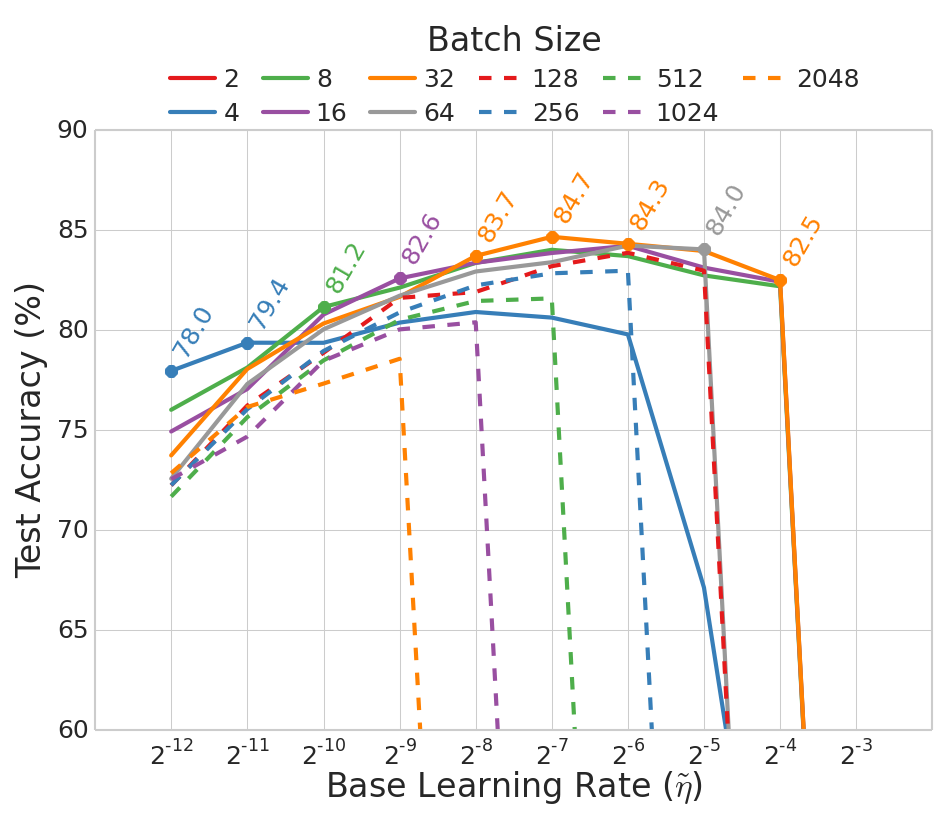}}
\end{subfigmatrix}
\caption{Test performance of reduced AlexNet model with BN, for different batch sizes and increasing values of $\tilde{\eta} = \eta/m$. CIFAR-10 dataset without data augmentation.}
\label{fig:AlexNet_CIFAR-10_BN}

\vspace{0.5cm}

\begin{subfigmatrix}{2}
\subfigure{\includegraphics[width=0.45\linewidth]{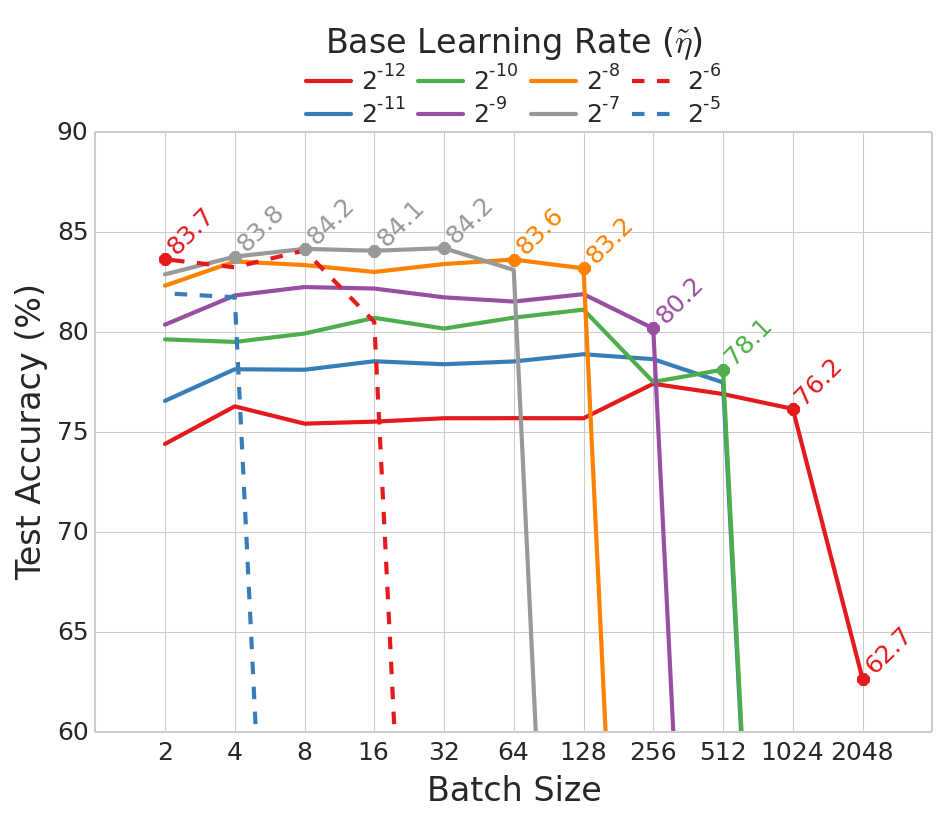}}
\subfigure{\includegraphics[width=0.45\linewidth]{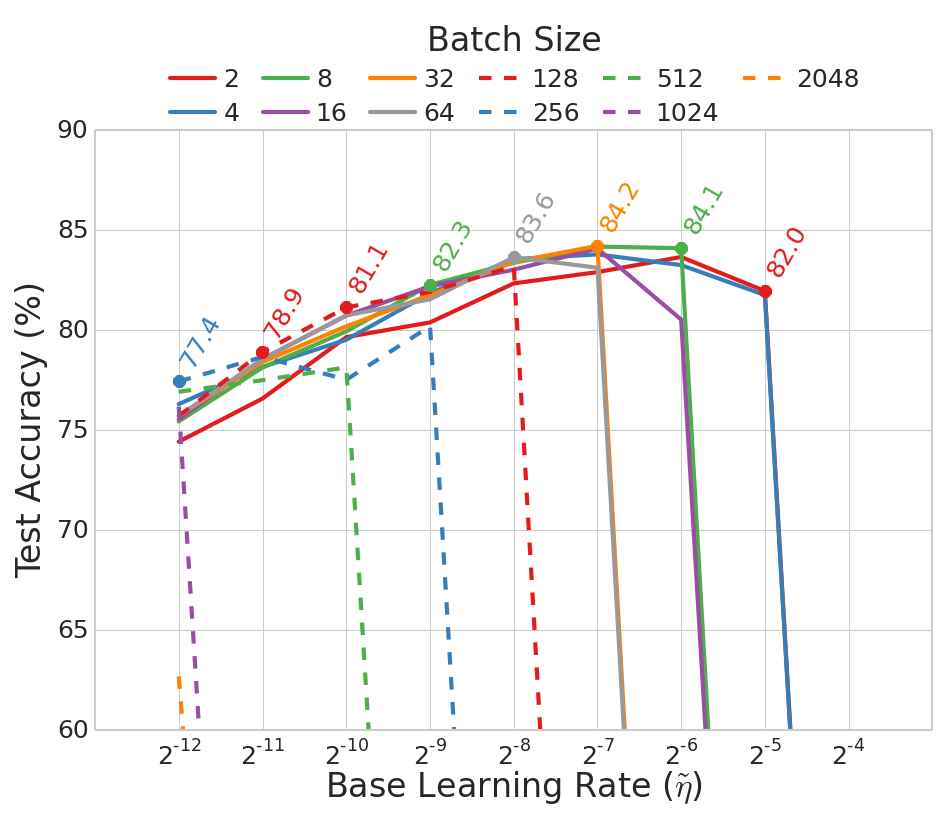}}
\end{subfigmatrix}
\caption{Test performance of reduced AlexNet model with BN only after the convolutions, for different batch sizes and increasing values of $\tilde{\eta} = \eta/m$. CIFAR-10 dataset without data augmentation.}
\label{fig:AlexNet_CIFAR-10_BN_conv_only}
\end{figure}

\begin{figure}[p]
\centering
\patchcmd{\subfigmatrix}{\hfill}{\hspace{0.6cm}}{}{} 
\begin{subfigmatrix}{2}
\subfigure{\includegraphics[width=0.45\linewidth]{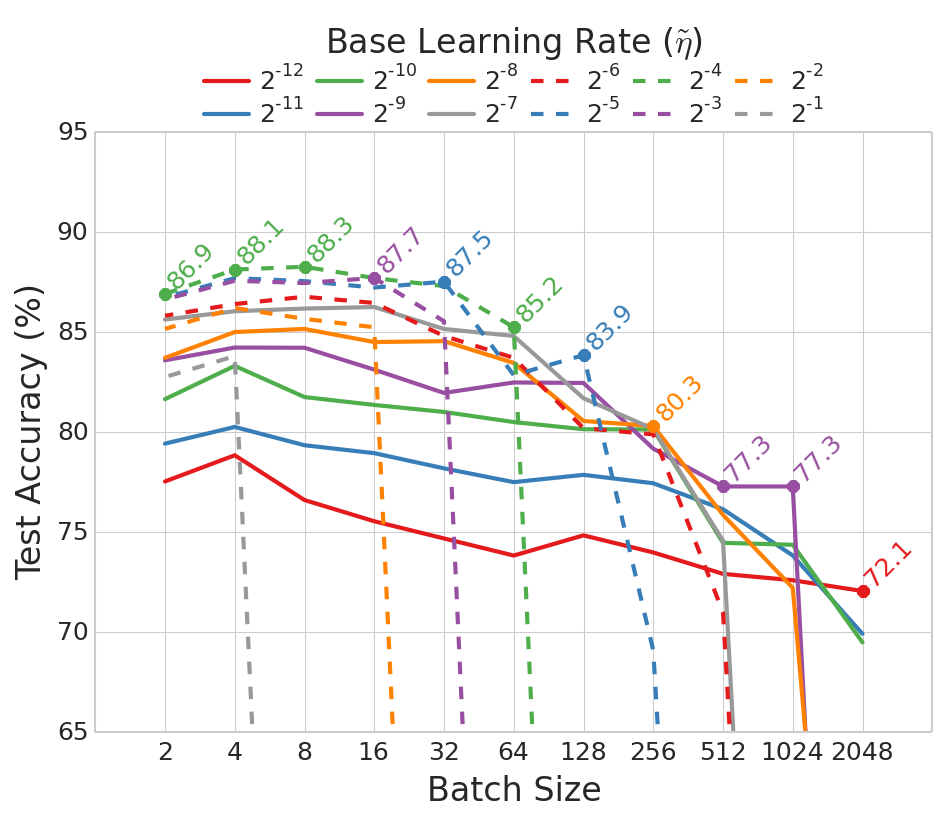}}
\subfigure{\includegraphics[width=0.45\linewidth]{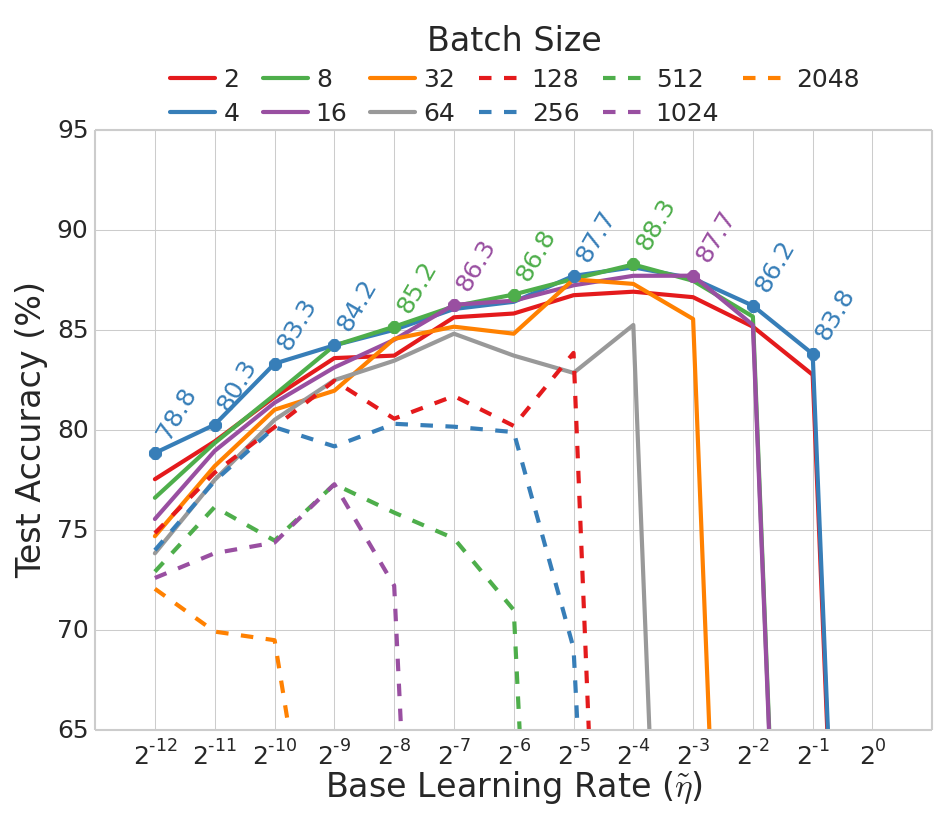}}
\end{subfigmatrix}
\caption{Test performance of ResNet-32 model with BN, for different batch sizes and increasing values of $\tilde{\eta} = \eta/m$. CIFAR-10 dataset without data augmentation.}
\label{fig:ResNet-32_CIFAR-10_noAug_BN}

\vspace{0.5cm}

\begin{subfigmatrix}{2}
\subfigure{\includegraphics[width=0.45\linewidth]{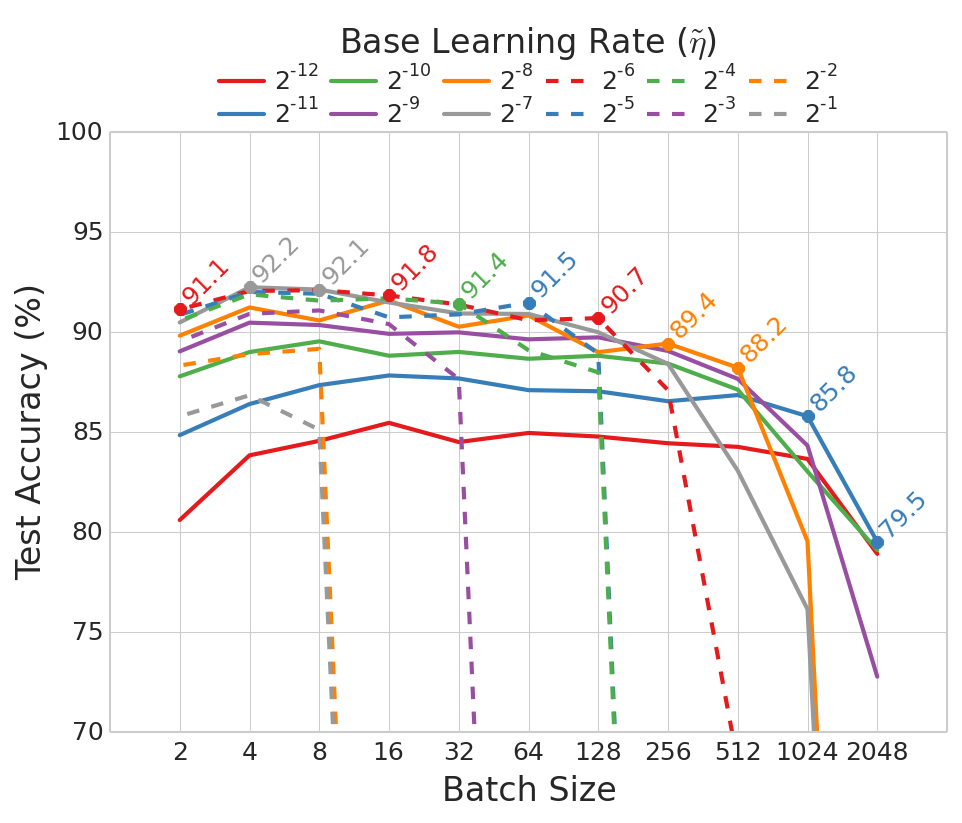}}
\subfigure{\includegraphics[width=0.45\linewidth]{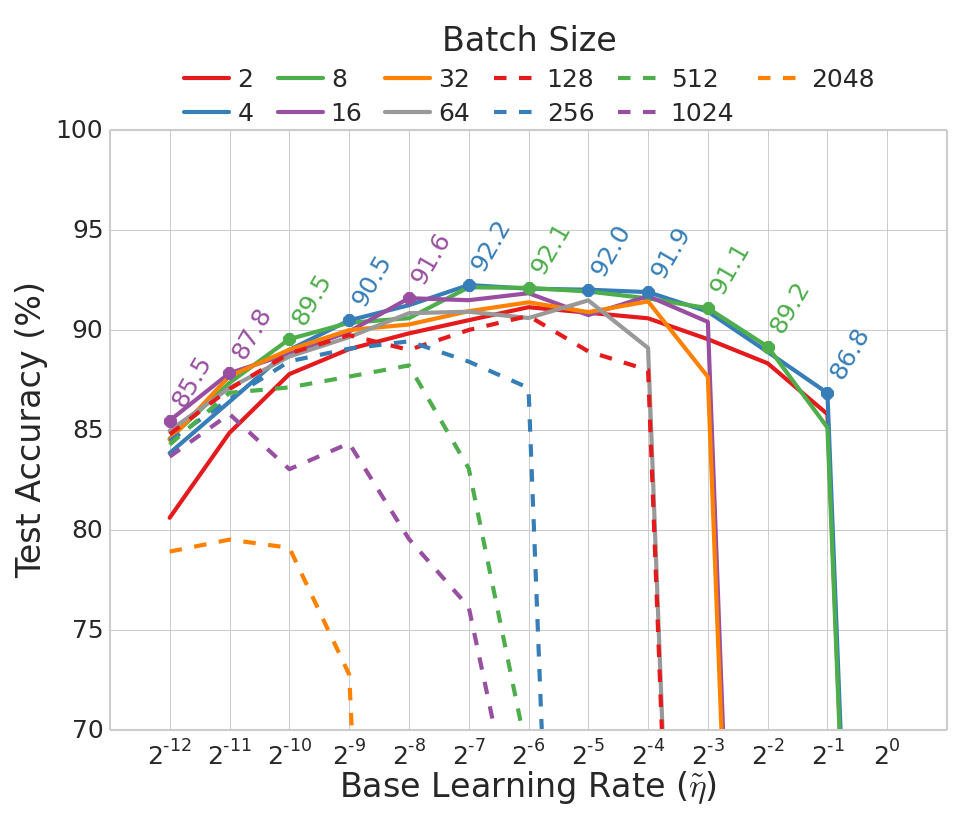}}
\end{subfigmatrix}
\caption{Test performance of ResNet-32 model with BN, for different batch sizes and increasing values of $\tilde{\eta} = \eta/m$. CIFAR-10 dataset with data augmentation.}
\label{fig:ResNet-32_CIFAR-10_Aug_BN}

\vspace{0.5cm}

\begin{subfigmatrix}{2}
\subfigure{\includegraphics[width=0.45\linewidth]{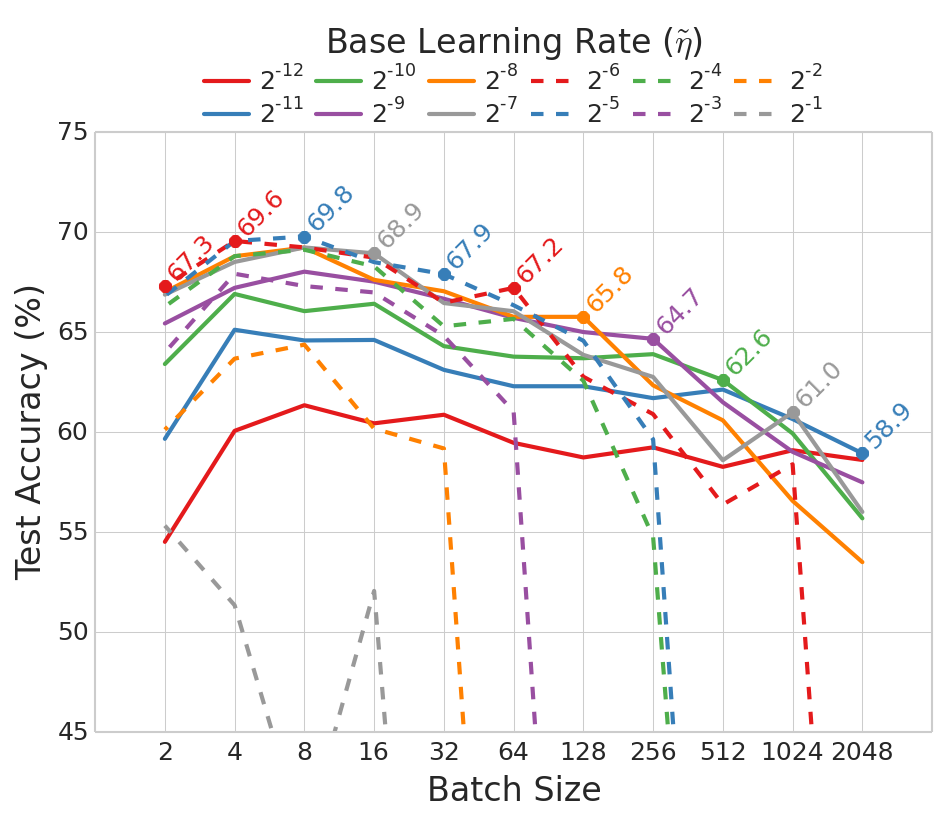}}
\subfigure{\includegraphics[width=0.45\linewidth]{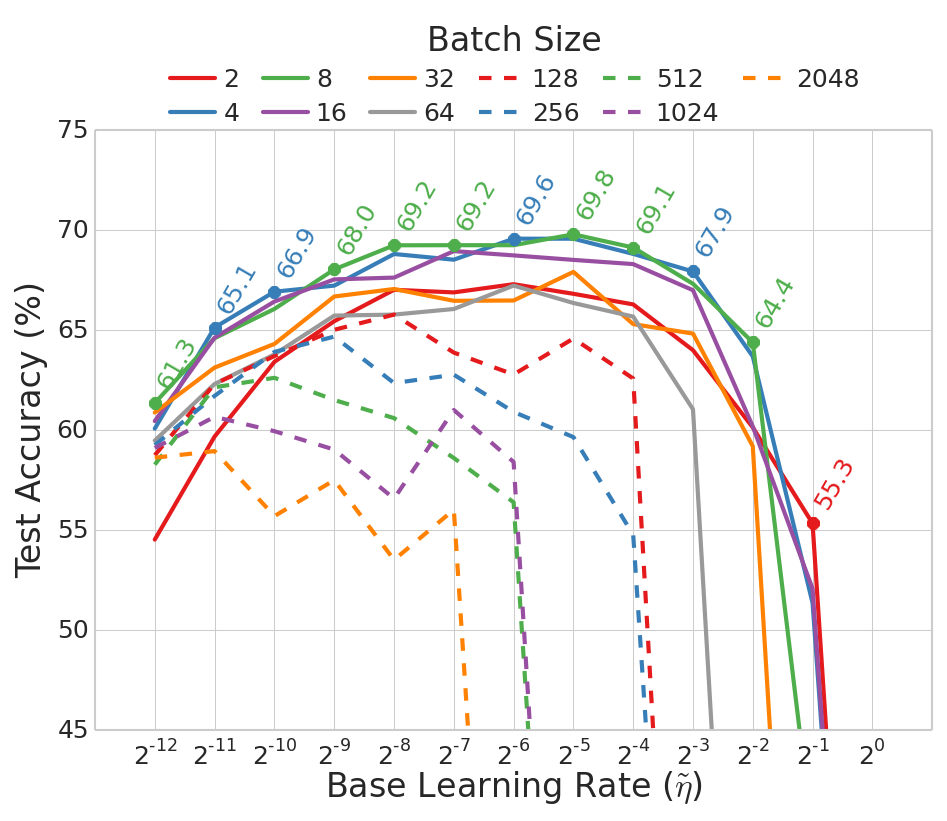}}
\end{subfigmatrix}
\caption{Test performance of ResNet-32 model with BN, for different batch sizes and increasing values of $\tilde{\eta} = \eta/m$. CIFAR-100 dataset with data augmentation.}
\label{fig:ResNet-32_CIFAR-100_Aug_BN}
\end{figure}

Figures~\ref{fig:AlexNet_CIFAR-10_BN} and~\ref{fig:AlexNet_CIFAR-10_BN_conv_only} show the CIFAR-10 test performance of the AlexNet model with BN but without data augmentation, for different values of batch size $m$ and base learning rate $\tilde{\eta}$.
BN is commonly applied to the activations in all convolutional and fully-connected layers. 
To assess the effect of BN on the different layers, here we have considered two different implementations of the reduced AlexNet model from Section~\ref{sec:Performance_noBN}: one with BN applied in both the convolutional and fully-connected layers, and a second with BN applied only in the convolutional layers.
The results show that the use of BN provides a consistent improvement in test accuracy for a wide range of batch sizes, compared to the results without BN presented in Figure~\ref{fig:AlexNet_CIFAR-10_noBN}.
As discussed in Section \ref{sec:Batch_Norm}, while the sample size for the estimation of the batch statistics does reduce with the batch size, for the convolutional layers the batch samples are also taken over the feature maps height and width. For the fully-connected layers, the batch sample size for the calculation of the mean and variance is instead simply equal to the batch size -- this can, unsurprisingly, cause issues for the smallest batch sizes investigated.
The use of BN for the fully-connected layers improves the convergence for larger batch sizes. However, the best final performance is still obtained for batch sizes between $m = 8$ and $m = 64$, compared with best batch sizes between $m = 4$ and $m = 32$ for the case where
BN is applied to only the convolutional layers (Figure~\ref{fig:AlexNet_CIFAR-10_BN_conv_only}). Both cases yield similar generalization performance.

Figures~\ref{fig:ResNet-32_CIFAR-10_noAug_BN}--\ref{fig:ResNet-32_CIFAR-100_Aug_BN} report the training performance for CIFAR-10 (with and without data augmentation) and CIFAR-100 (with data augmentation) on the ResNet-32 model with BN, for different values of batch size $m$ and base learning rate $\tilde{\eta}$.
All results show a significant performance degradation for increasing values of the batch size, with the best results obtained for batch sizes $m = 4$ or $m = 8$.

The results reported in Figures~\ref{fig:ResNet-32_CIFAR-10_noAug_BN}--\ref{fig:ResNet-32_CIFAR-100_Aug_BN} indicate in each of the different cases a clear optimum value of base learning rate, which is only achievable for batch sizes $m = 16$ or smaller for CIFAR-10, and $m=8$ or smaller for CIFAR-100.

\subsection{Performance With Gradual Warm-up}
\label{sec:Performance_BN_WU}

As discussed in Section \ref{sec:Other_Work}, warm-up strategies have previously been employed to improve the performance of large batch training~\citep{Goyal17}. The results reported in~\citet{Goyal17} have shown that the ImageNet validation accuracy for ResNet-50 for batch size $m = 256$  could be matched over the same number of epochs with batch sizes of up to $8192$, by applying a suitable gradual warm-up procedure. 
In our experiments, the same warm-up strategy has been used to investigate a possible similar improvement for the case of small batch sizes.

\begin{figure}[p]
\centering
\patchcmd{\subfigmatrix}{\hfill}{\hspace{0.6cm}}{}{} 
\begin{subfigmatrix}{2}
\subfigure{\includegraphics[width=0.45\linewidth]{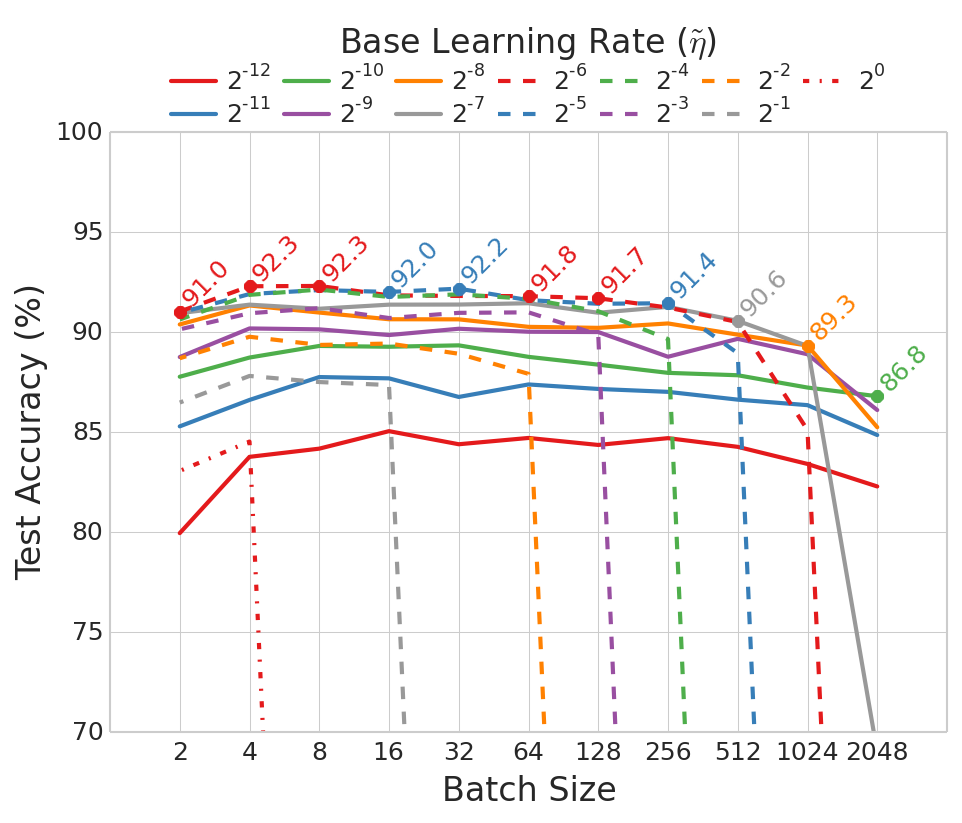}}
\subfigure{\includegraphics[width=0.45\linewidth]{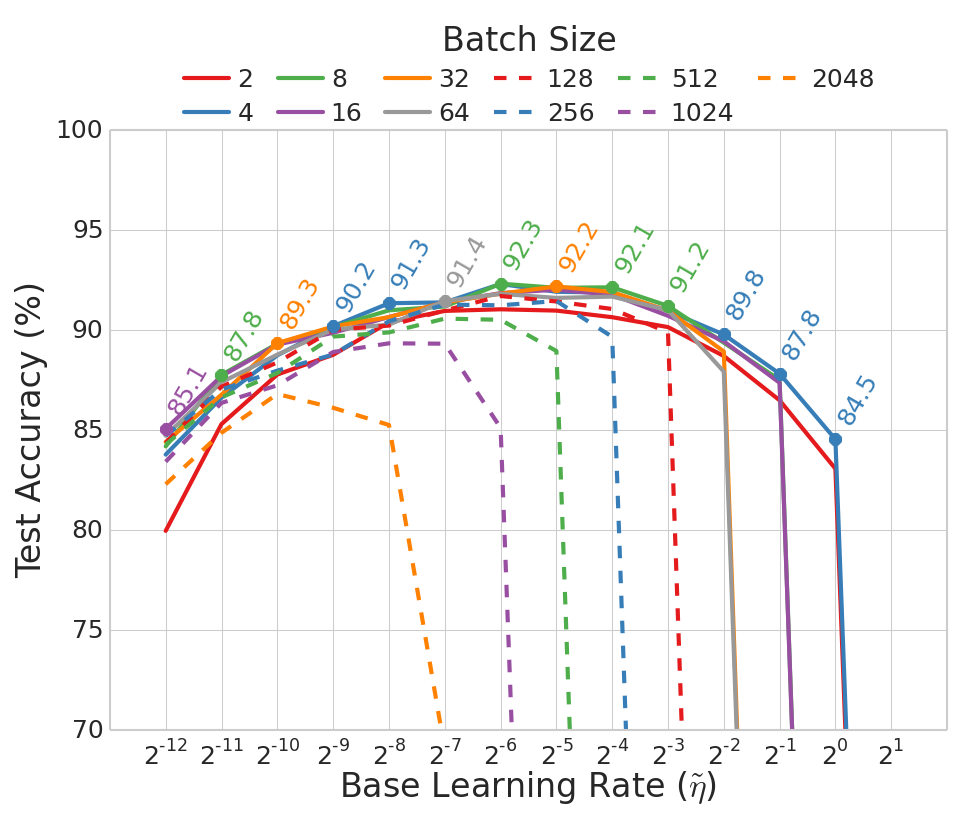}}
\end{subfigmatrix}
\caption{Test performance of ResNet-32 model with BN and with the warm-up strategy of~\citet{Goyal17}. CIFAR-10 dataset with data augmentation.}
\label{fig:ResNet-32_CIFAR-10_WU_Aug_BN}

\vspace{0.5cm}

\begin{subfigmatrix}{2}
\subfigure{\includegraphics[width=0.45\linewidth]{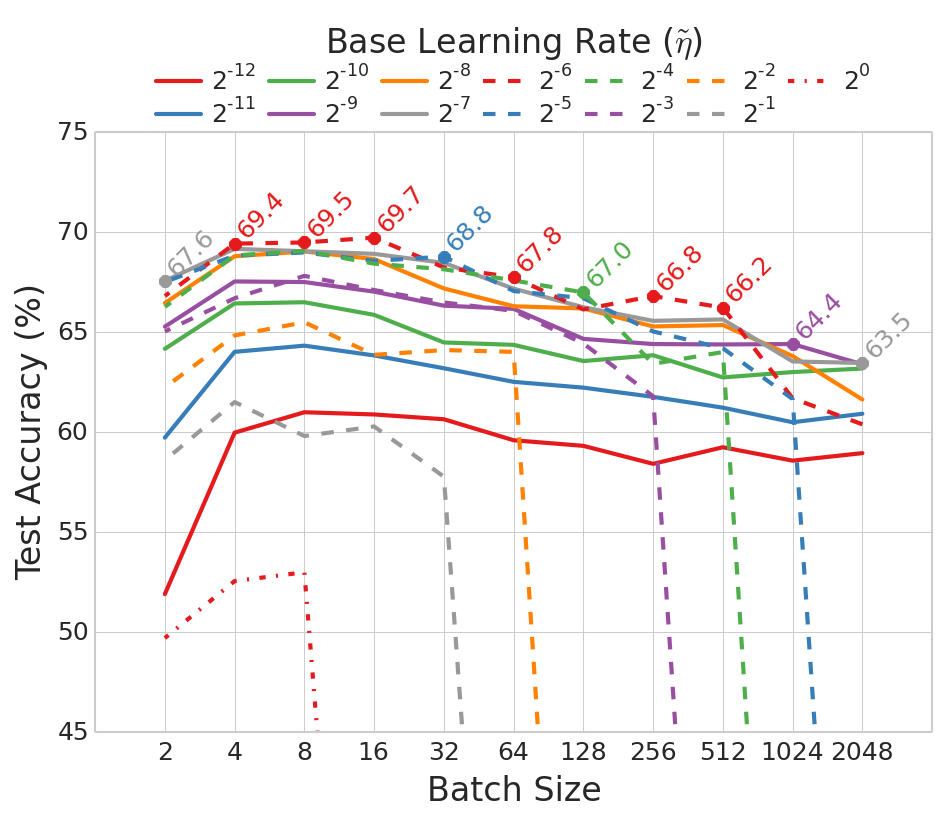}}
\subfigure{\includegraphics[width=0.45\linewidth]{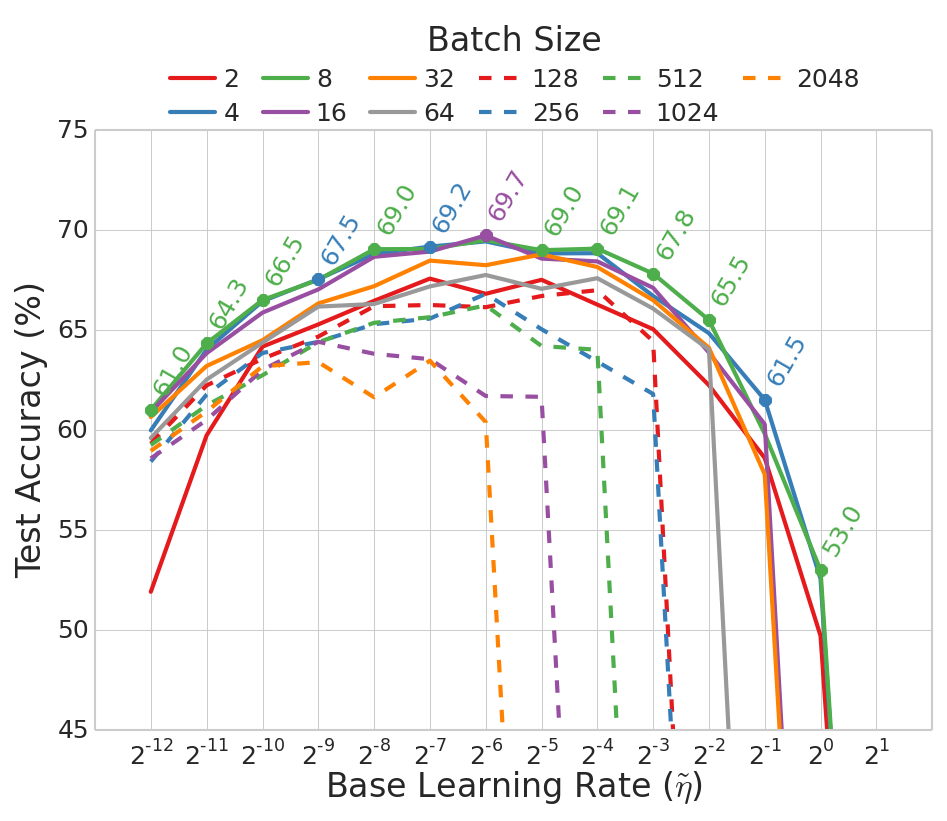}}
\end{subfigmatrix}
\caption{Test performance of ResNet-32 model with BN and with the warm-up strategy of~\citet{Goyal17}. CIFAR-100 dataset with data augmentation.}
\label{fig:ResNet-32_CIFAR-100_WU_Aug_BN}

\vspace{0.5cm}

\begin{subfigmatrix}{2}
\subfigure{\includegraphics[width=0.45\linewidth]{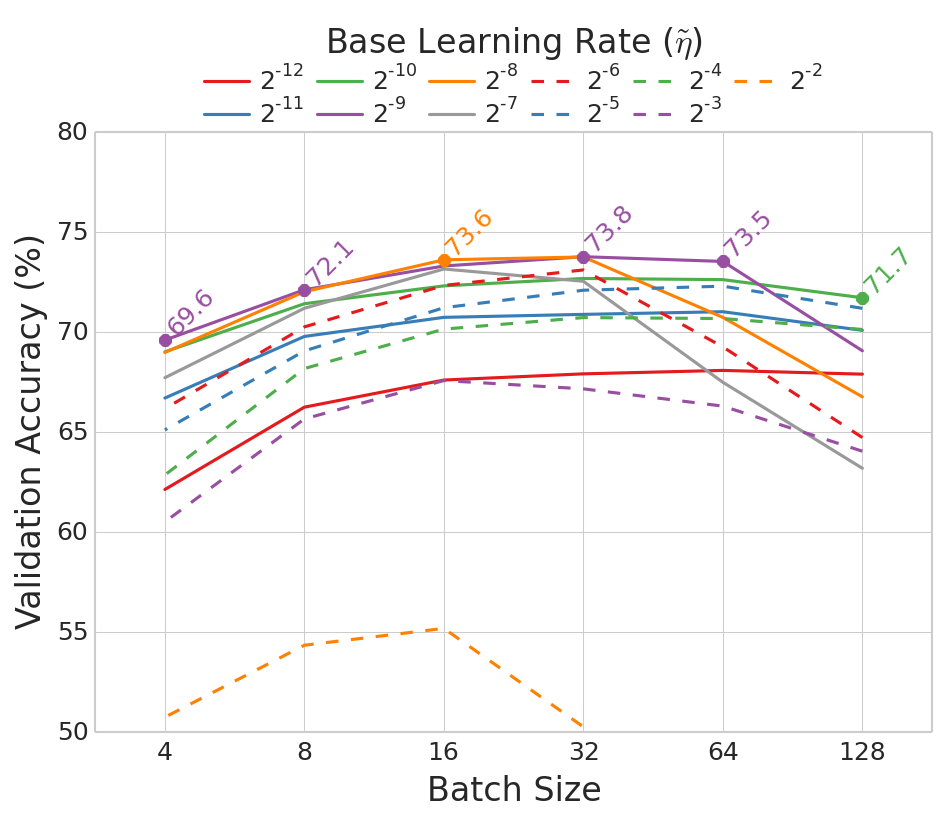}}
\subfigure{\includegraphics[width=0.45\linewidth]{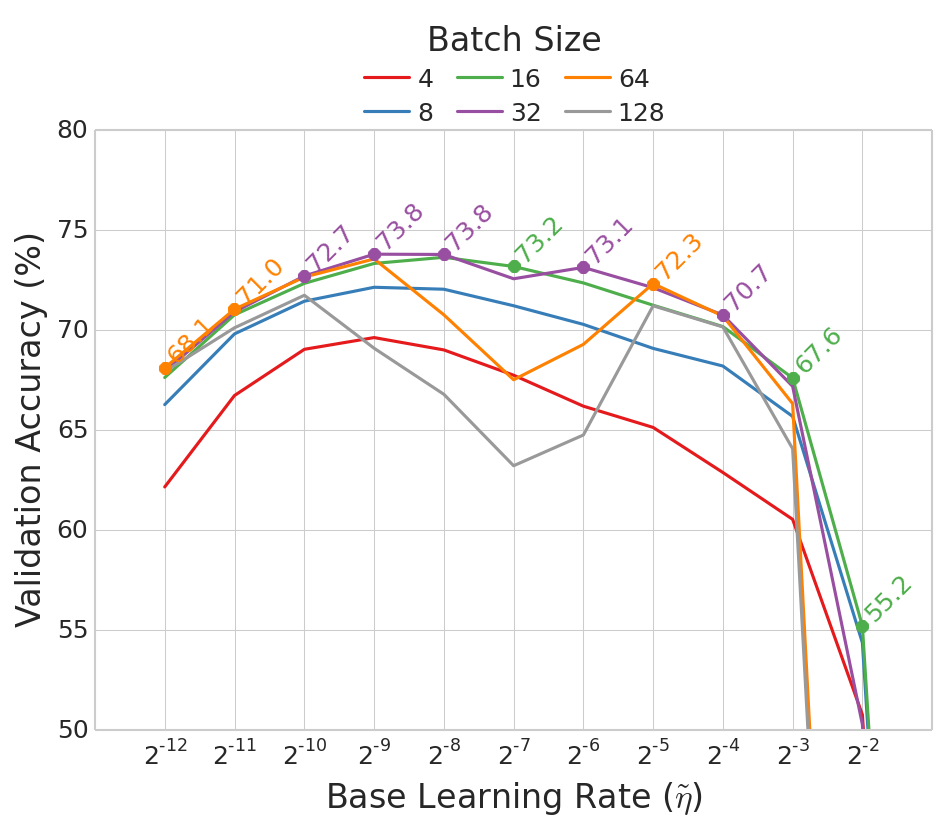}}
\end{subfigmatrix}
\caption{Top-1 validation performance of ResNet-50 model with BN, for different batch sizes and increasing values of $\tilde{\eta} = \eta/m$. ImageNet dataset with data augmentation. The reported results are the average of two runs.}
\label{fig:ResNet-50_ImageNet_Top1}
\end{figure}

Following the procedure of~\citet{Goyal17}, the implementation of the gradual warm-up corresponds to the use of an initial learning rate of $\eta/32$, with a linear increase from $\eta/32$ to $\eta$ over the first 5\% of training. The standard learning rate schedule is then used for the rest of training.
Here, the above gradual warm-up has been applied to the training of the ResNet-32 model, for both CIFAR-10 and CIFAR-100 datasets with BN and data augmentation.
The corresponding performance results are reported in Figures~\ref{fig:ResNet-32_CIFAR-10_WU_Aug_BN} and \ref{fig:ResNet-32_CIFAR-100_WU_Aug_BN}.

As discussed in Section~\ref{sec:Other_Work}, the gradual warm-up helps to maintain consistent training dynamics in the early stages of optimization, where the approximation (\ref{eq:SGD_6}) is expected to be particularly weak.
From Figures~\ref{fig:ResNet-32_CIFAR-10_WU_Aug_BN} and \ref{fig:ResNet-32_CIFAR-100_WU_Aug_BN}, the use of the gradual warm-up strategy proposed in~\citet{Goyal17} generally improves the training performance. However, the results also show that the best performance is still obtained for smaller batch sizes. For the CIFAR-10 results (Figure~\ref{fig:ResNet-32_CIFAR-10_WU_Aug_BN}), the best test accuracy corresponds to batch sizes $m = 4$ and $m = 8$, although quite good results are maintained out to $m=128$, while for CIFAR-100 (Figure~\ref{fig:ResNet-32_CIFAR-100_WU_Aug_BN}) the best test performance is obtained for batch sizes between $m = 4$ and $m = 16$, with a steeper degradation for larger batch sizes.

\subsection{ImageNet Performance}
\label{sec:ImageNet_Performance}

The ResNet-50 performance for the ImageNet dataset is reported in Figure~\ref{fig:ResNet-50_ImageNet_Top1}, confirming again that the best performance is achieved with smaller batches. The best validation accuracy is obtained with batch sizes between $m = 16$ and $m = 64$.

The ImageNet results also show less predictable behaviour for larger batch sizes. In contrast to the CIFAR-10 and CIFAR-100 results, the performance degradation with larger batch sizes is observed to be particularly severe over a specific range of base learning rates $\tilde{\eta}$.
Batch sizes $m = 16$ and $m = 32$ achieve the best performance over a continuous range of base learning rates, while the performance with larger batch sizes degrades precisely over the range of base learning rates that corresponds to the best overall performance.
On the other hand, smaller batch sizes have shown stable and consistent convergence over the full range of learning rates.

Overall, the ImageNet results confirm that the use of small batch sizes provides performance advantages and reliable training convergence for an increased range of learning rates.

\subsection{Different Batch Size for Weight Update and Batch Normalization}
\label{sec:Dist_Performance}

\begin{figure}[tb]
\patchcmd{\subfigmatrix}{\hfill}{\hspace{0.6cm}}{}{} 
\begin{subfigmatrix}{2}
\subfigure[CIFAR-10]{\includegraphics[width=0.45\linewidth]{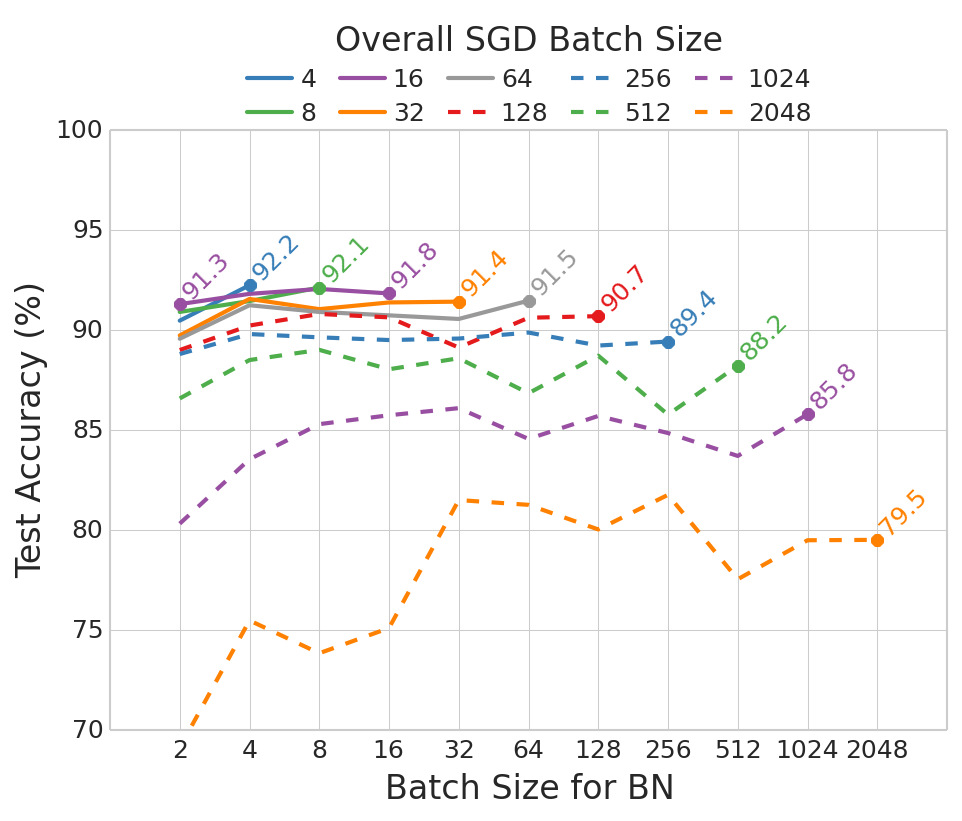}
\label{fig:ResNet32_CIFAR10_Aug_VB}}
\subfigure[CIFAR-100]{\includegraphics[width=0.45\linewidth]{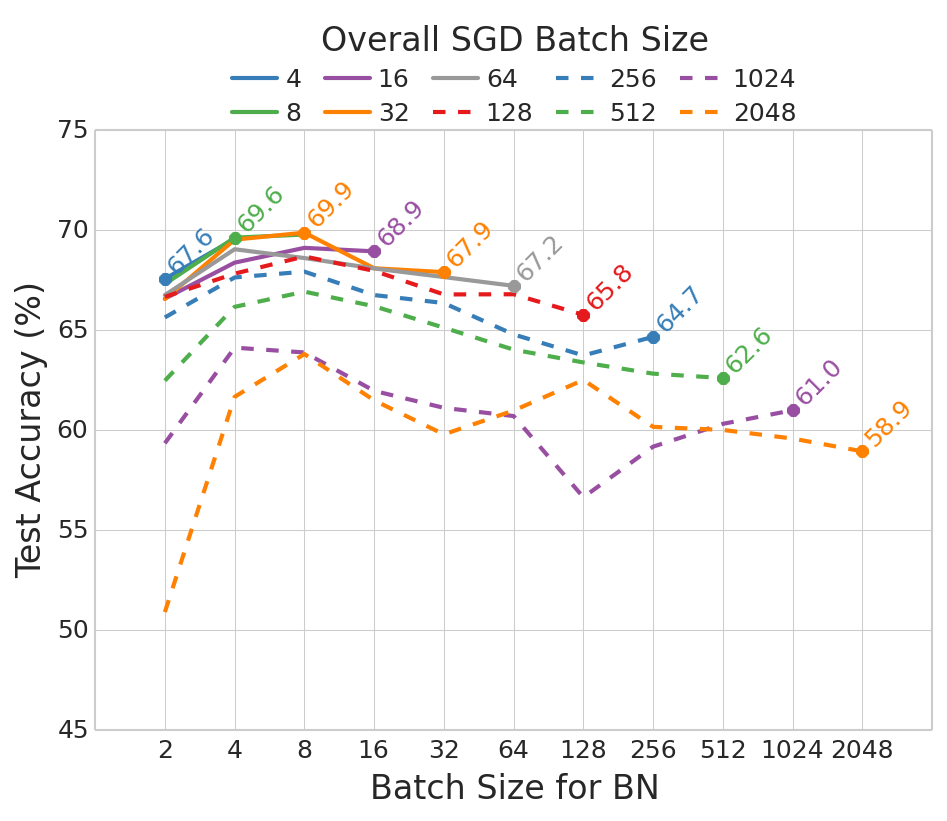}
\label{fig:ResNet32_CIFAR100_Aug_VB}}
\end{subfigmatrix}
\caption{Test performance of ResNet-32 model with BN, for different values of the overall SGD batch size and the batch size for BN.  CIFAR-10 and CIFAR-100 datasets with data augmentation.}
\label{fig:ResNet32_CIFAR_Aug_VB}

\vspace{0.5cm}

\centering
\includegraphics[width=0.45\textwidth]{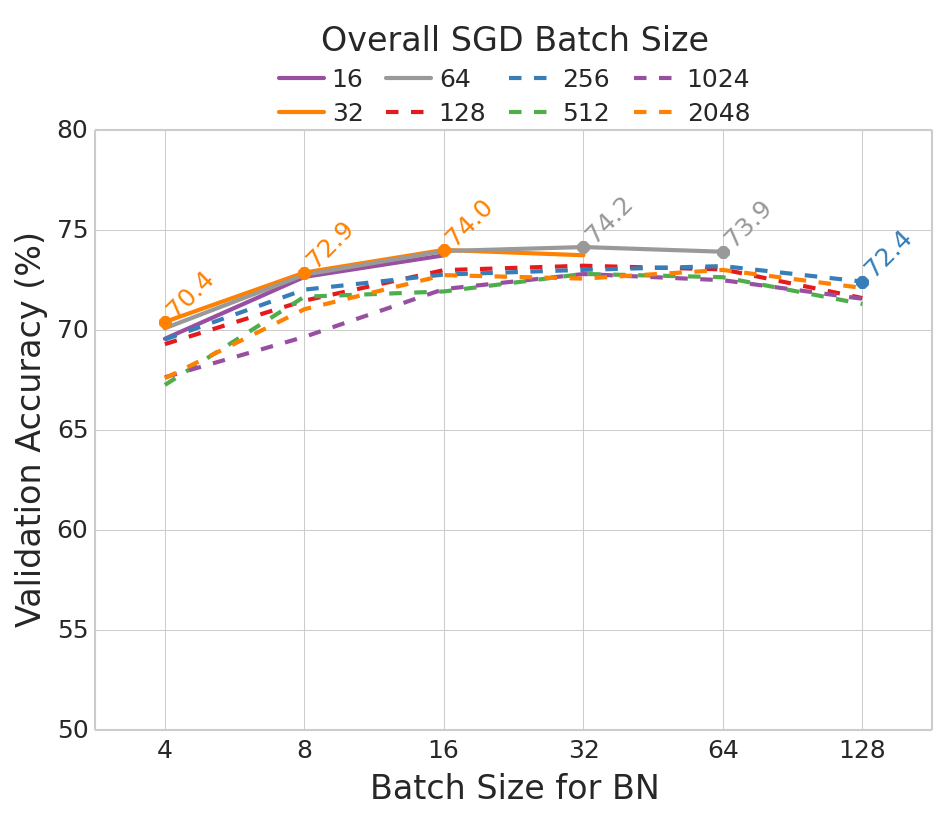}
\caption{Top-1 validation performance of ResNet-50 model with BN, for different values of the overall SGD batch size and the batch size for BN. ImageNet dataset with data augmentation.}
\label{fig:ResNet50_VB}
\end{figure}

In the experimental results presented in Sections~\ref{sec:Performance_noBN} to~\ref{sec:ImageNet_Performance} (Figures~\ref{fig:AlexNet_CIFAR-10_noBN}--\ref{fig:ResNet-50_ImageNet_Top1}), the same batch size has been used both for the SGD weight updates and for BN. We now consider the effect of using small sub-batches for BN, and  larger batches for SGD. This is common practice for the case of data parallel distributed processing, where BN is often implemented independently on each individual worker, while the overall batch size for the SGD weight updates is the aggregate of the BN sub-batches across all workers.

Figure~\ref{fig:ResNet32_CIFAR_Aug_VB} shows the CIFAR-10 and CIFAR-100 results for different values of the overall SGD batch size and of the batch size for BN.
Each curve corresponds to a fixed value of overall SGD batch size, where the experiments have been run using the base learning rate $\tilde{\eta}$ which delivered the best results in Section~\ref{sec:Performance_BN} (Figures~\ref{fig:ResNet-32_CIFAR-10_Aug_BN} and \ref{fig:ResNet-32_CIFAR-100_Aug_BN}) for the same SGD batch size.

The results reported in Figure~\ref{fig:ResNet32_CIFAR_Aug_VB} show a general performance improvement by reducing the overall batch size for the SGD weight updates, in line with the analysis of Section~\ref{sec:Background}. We also note that, for the CIFAR-100 results, the best test accuracy for a given overall batch size is consistently obtained when even smaller batches are used for BN. This is against the intuition that BN should benefit from estimating the normalization statistics over larger batches. This evidence suggests that, for a given overall batch size and fixed base learning rate, the best values of the batch size for BN are typically smaller that the batch size used for SGD. Moreover, the results indicate that the best values of the batch size for BN are only weakly related to the SGD batch size, possibly even independent of it. For example, a BN batch size $m = 4$ or $m = 8$ appears to give best results for all SGD batch sizes tested.

Figure~\ref{fig:ResNet50_VB} reports the corresponding ImageNet results for different values of the overall batch size for the SGD weight updates and of the batch size for BN. Again, each curve is for a fixed value of the SGD batch size, using the base learning rate $\tilde{\eta}$ which yielded the best results in Figure~\ref{fig:ResNet-50_ImageNet_Top1} with that batch size. For overall SGD batch sizes larger than $128$, the best base learning rate has been chosen based on experiments with BN performed over batches of $128$; the  selected base learning rates were $\tilde{\eta} = 10^{-10}$ for overall batch sizes $256$, $512$ and $1024$ and $\tilde{\eta} = 10^{-9}$ for overall batch size $2048$.

The results presented in Figure~\ref{fig:ResNet50_VB} show that the performance with BN generally improves by reducing the overall batch size for the SGD weight updates. The collected data supports the conclusion that the possibility of achieving the best test performance depends on the use of a small batch size for the overall SGD optimization.

The plots also indicate that the best performance is obtained when smaller batches are used for BN in the range between $16$ and $64$, for all SGD batch sizes tested. From these results, the advantages of using a small batch size for both SGD weight updates and BN suggests that the best solution for a distributed implementation would be to use a small batch size per worker, and distribute \textit{both} BN and SGD over multiple workers.

\section{Discussion}
\label{sec:Discussion}

The reported experiments have explored the training dynamics and generalization performance of small batch training for different datasets and neural networks.

\begin{figure}[tb]
\centering
\includegraphics[width=0.75\textwidth]{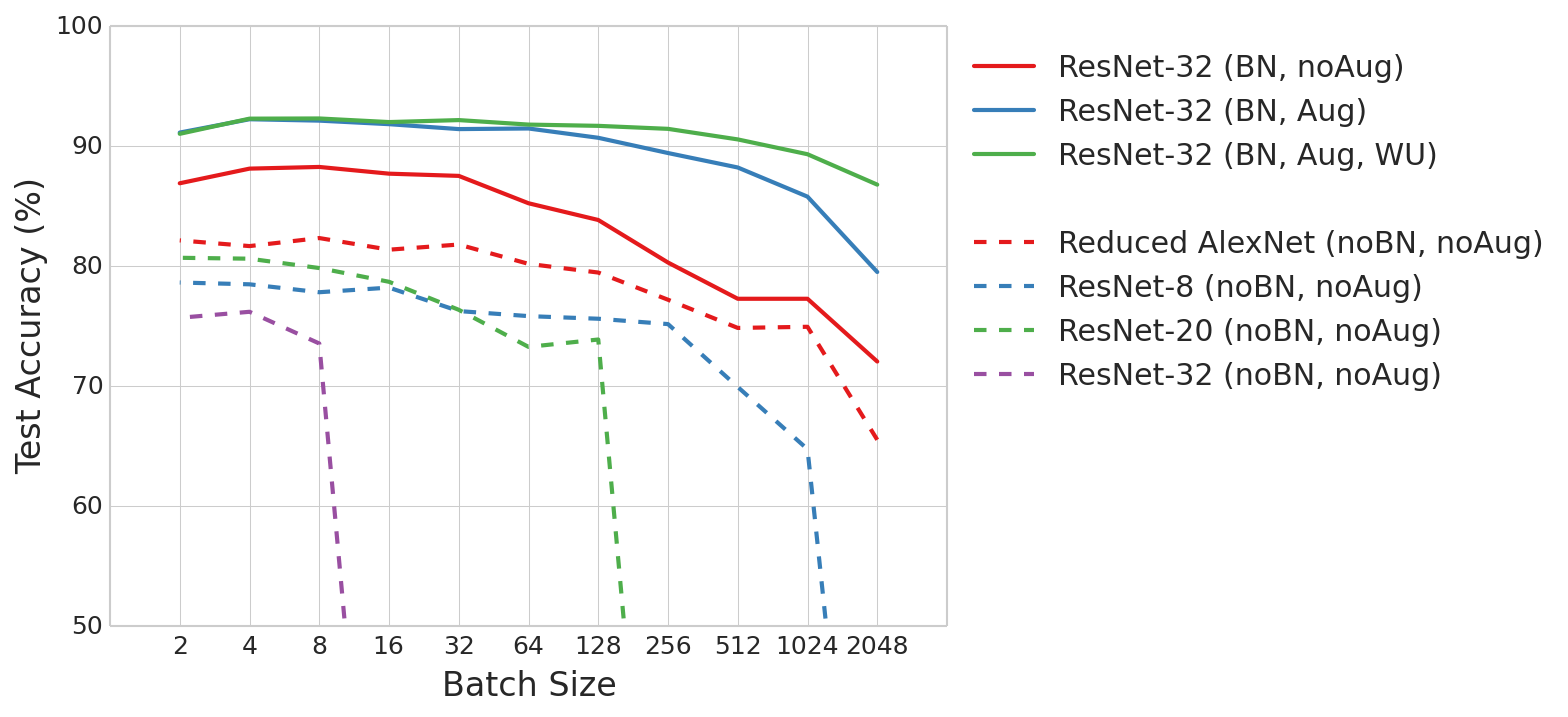}
\caption{Summary of the best CIFAR-10 test performance achieved with different batch sizes, for the different network architectures presented in Section~\ref{sec:Batch_Train_Performance}. (\texttt{BN}, \texttt{noBN}: with/without BN; \texttt{Aug}, \texttt{noAug}: with/without data augmentation; \texttt{WU}: with gradual warmup.)}
\label{fig:Val_Summary}
\end{figure}

Figure~\ref{fig:Val_Summary} summarizes the best CIFAR-10 results achieved for a selection of the network architectures discussed in Sections \ref{sec:Batch_Train_Performance}. The curves show that the best test performance is always achieved with a batch size below $m = 32$. In the cases without BN, the performance is seen to continue to improve down to batch size $m = 2$. This is in line with the analysis of Section~\ref{sec:Perspective}, that suggests that, with smaller batch sizes, the possibility of using more up-to-date gradient information always has a positive impact on training.
The operation with BN is affected by the issues highlighted in Section \ref{sec:Batch_Norm}. However, the performance with BN appears to be maintained for batch sizes even as small as $m = 4$ and $m = 8$.

\begin{figure}[tb]
\centering
\includegraphics[width=1\textwidth]{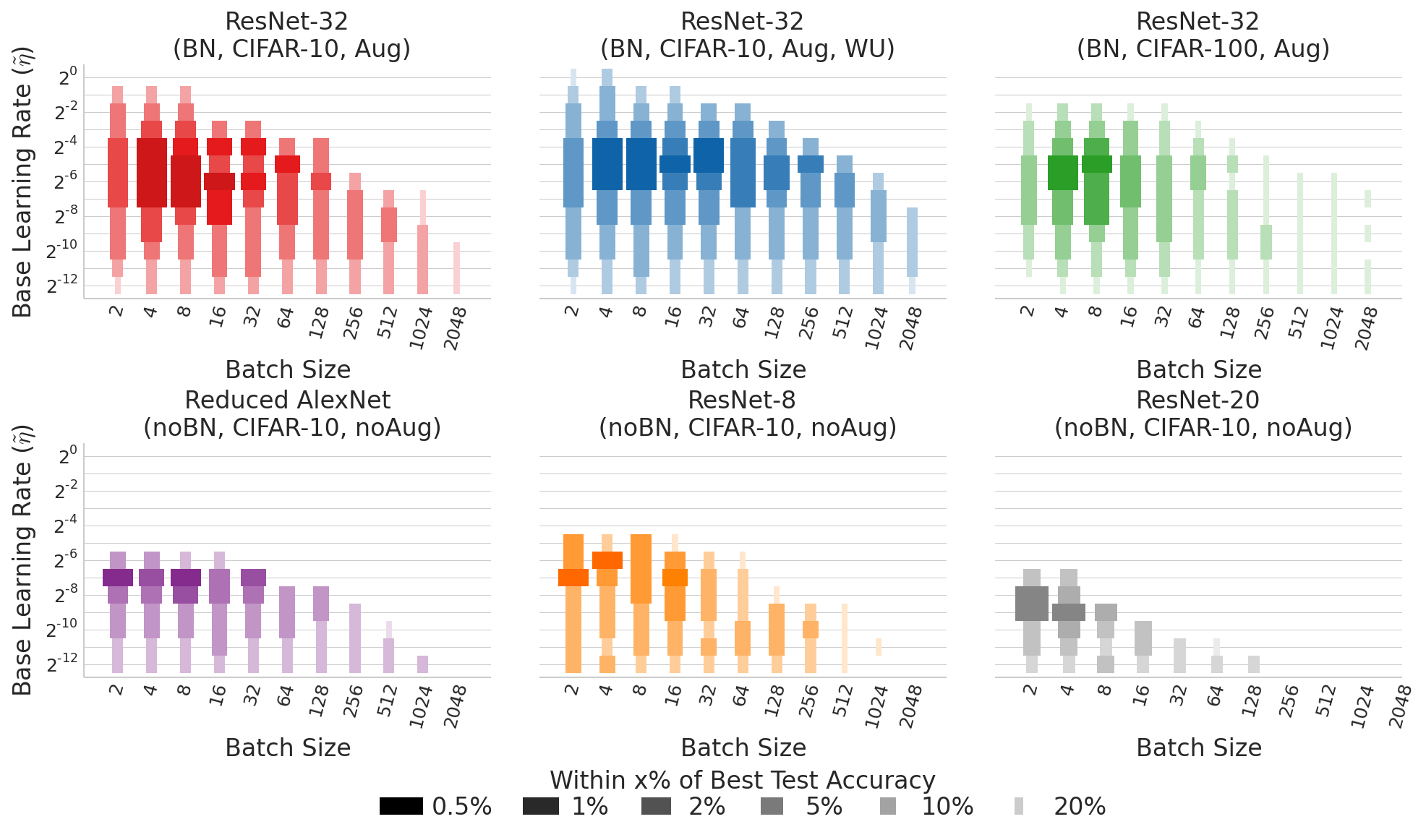}
\caption{Summary of range of base learning rates $\tilde{\eta} = \eta/m$ that provide reliable convergence, for different network architectures. CIFAR-10 and CIFAR-100 datasets. (\texttt{BN}, \texttt{noBN}: with/without BN; \texttt{Aug}, \texttt{noAug}: with/without data augmentation; \texttt{WU}: with gradual warmup.)}
\label{fig:LR_Summary}
\end{figure}

For the more difficult ImageNet training, Figure~\ref{fig:ResNet-50_ImageNet_Top1} shows that a similar trend can be identified. The best performance is achieved with batch sizes between $m = 16$ and $m = 64$, although for $m=64$ SGD becomes acutely sensitive to the choice of learning rate. The performance for batch sizes $m \leq 8$ depends on the discussed effect of BN. As mentioned in Section~\ref{sec:Batch_Norm}, the issue of the BN performance for very small batch sizes can be addressed by adopting alternative normalization techniques like Group Normalization~\citep{Wu18}.

\citet{Goyal17} have shown that the accuracy of ResNet-50 training on ImageNet for batch size $256$ could be maintained with batch sizes of up to $8192$ by using a gradual warm-up scheme, and with a BN batch size of $32$ whatever the SGD batch size. This warm-up strategy has been also applied to the CIFAR-10 and CIFAR-100 cases investigated here.
In our experiments, the use of a gradual warm-up did improve the performance of the large batch sizes, but did not fully recover the performance achieved with the smallest batch sizes.

Figure~\ref{fig:LR_Summary} gives an alternative view of the AlexNet and ResNet results presented in Section~\ref{sec:Batch_Train_Performance}, showing the range of base learning rates $\tilde{\eta} = \eta/m$ that provide reliable convergence for each batch size.
The results show that increasing the batch size progressively reduces the range of learning rates that provide stable convergence. This demonstrates how the increased variance in the weight update associated with the use of larger batch sizes can affect the robustness and stability of training, and highlights the limitations of the linear scaling rule.
Furthermore, the range of results reported in Section~\ref{sec:Batch_Train_Performance} shows that, for each experiment, there is a clear optimal base learning rate $\tilde{\eta}$, but stable convergence can often only be achieved using this value of $\tilde{\eta}$ in combination with a small batch size. This suggests that performing a sweep of the learning rate for large batch sizes could easily lead to an incorrect conclusion on the optimal learning rate.

While the presented results advocate the use of small batch sizes to improve the convergence and accuracy, the use of small batch sizes also has the effect of reducing the computational parallelism available. This motivates the need to consider the trade-off between hardware efficiency and test performance.

We should also observe that the experiments performed in this paper refer to some of the most successful deep network architectures that have been recently proposed. 
These models have typically been designed under the presumption that large batches are necessary for good machine performance. The realisation that small batch training is preferable for both generalization performance and training stability should motivate re-examination of machine architectural choices which perform well with small batches; there exists at least the opportunity to exploit the small memory footprint of small batches. 
Moreover, it can be expected that small batch training may provide further benefits for model architectures that have inherent stability issues.

We have also investigated the performance with different batch sizes for SGD and BN, which has often been used to reduce communication in a distributed processing setting. The evidence presented in this work shows that in these circumstances the largest contributor to the performance was the value of the overall batch size for SGD.
For the overall SGD batch sizes providing the best test accuracy, the best performance was achieved using values of batch size for BN between $m = 4$ and $m = 8$ for the CIFAR-10 and CIFAR-100 datasets, or between $m = 16$ and $m = 64$ for the ImageNet dataset.
These results further confirm the need to consider the trade-off between accuracy and efficiency, based on the fact that to achieve the best test performance it is important to use a small overall batch size for SGD. In order to widely distribute the work, this implies 
that the best solution would be to distribute the implementation of both BN and SGD optimization over multiple workers.

\section{Conclusions}
\label{sec:Conclusions}
We have presented an empirical study of the performance of mini-batch stochastic gradient descent, and reviewed the underlying theoretical assumptions relating training duration and learning rate scaling to mini-batch size.

The presented results confirm that using small batch sizes achieves the best training stability and generalization performance, for a given computational cost, across a wide range of experiments. In all cases the best results have been obtained with batch sizes $m = 32$ or smaller, often as small as $m = 2$ or $m = 4$.

With BN and larger datasets, larger batch sizes can be useful, up to batch size $m = 32$ or $m = 64$.
However, these datasets would typically require a distributed implementation to avoid excessively long training. In these cases, the best solution would be to implement both BN and stochastic gradient optimization over multiple processors, which would imply the use of a small batch size per worker. We have also observed that the best values of the batch size for BN are often smaller than the overall SGD batch size.

The results  also highlight the optimization difficulties associated with large batch sizes.  The range of usable base learning rates significantly decreases for larger batch sizes, often to the extent that the optimal learning rate could not be used. We suggest that this can be attributed to a linear increase in the variance of the weight updates with the batch size.

Overall, the experimental results support the broad conclusion that using small batch sizes for training provides benefits both in terms of range of learning rates that provide stable convergence and achieved test performance for a given number of epochs.

\vspace{0.5cm}

\bibliographystyle{iclr2017_conference}
\bibliography{small_batch_training}

\vspace{0.5cm}

\appendix

\section{Additional Results with Batch Normalization}

\begin{figure}[htbp]
\centering
\patchcmd{\subfigmatrix}{\hfill}{\hspace{0.6cm}}{}{} 
\begin{subfigmatrix}{2}
\subfigure{\includegraphics[width=0.45\linewidth]{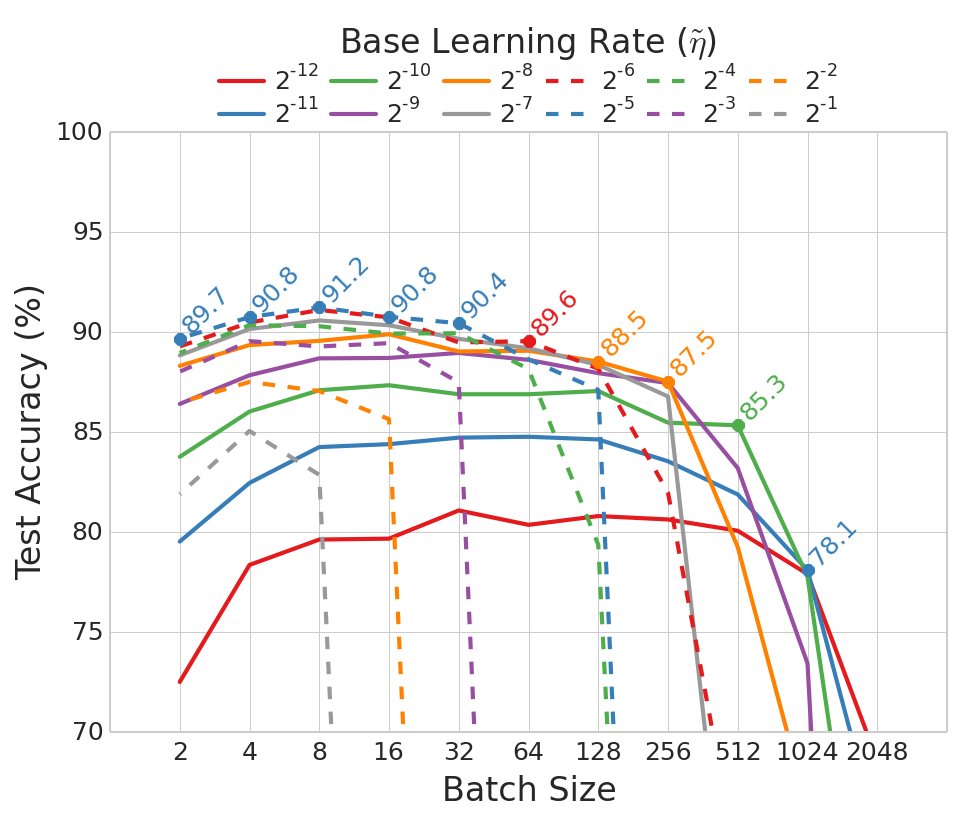}}
\subfigure{\includegraphics[width=0.45\linewidth]{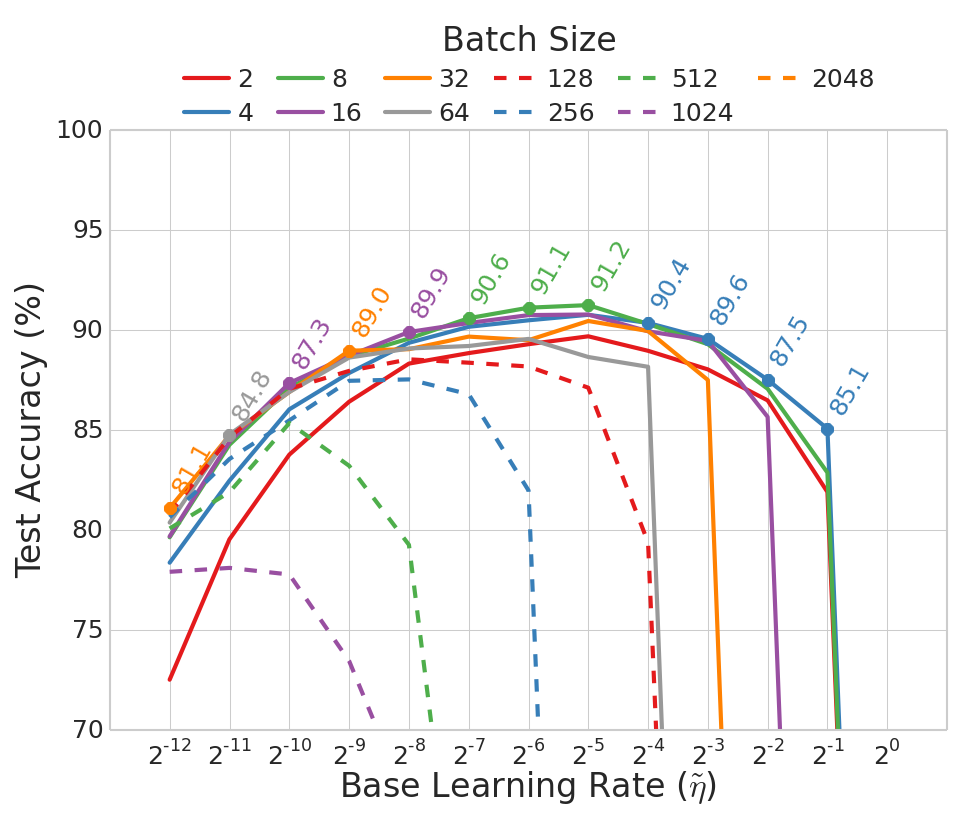}}
\end{subfigmatrix}
\caption{Test performance of ResNet-32 model with BN, for reduced training length in number of epochs. CIFAR-10 dataset with data augmentation.}
\label{fig:ResNet-32_CIFAR-10_Aug_BN_05}

\vspace{0.5cm}

\begin{subfigmatrix}{2}
\subfigure{\includegraphics[width=0.45\linewidth]{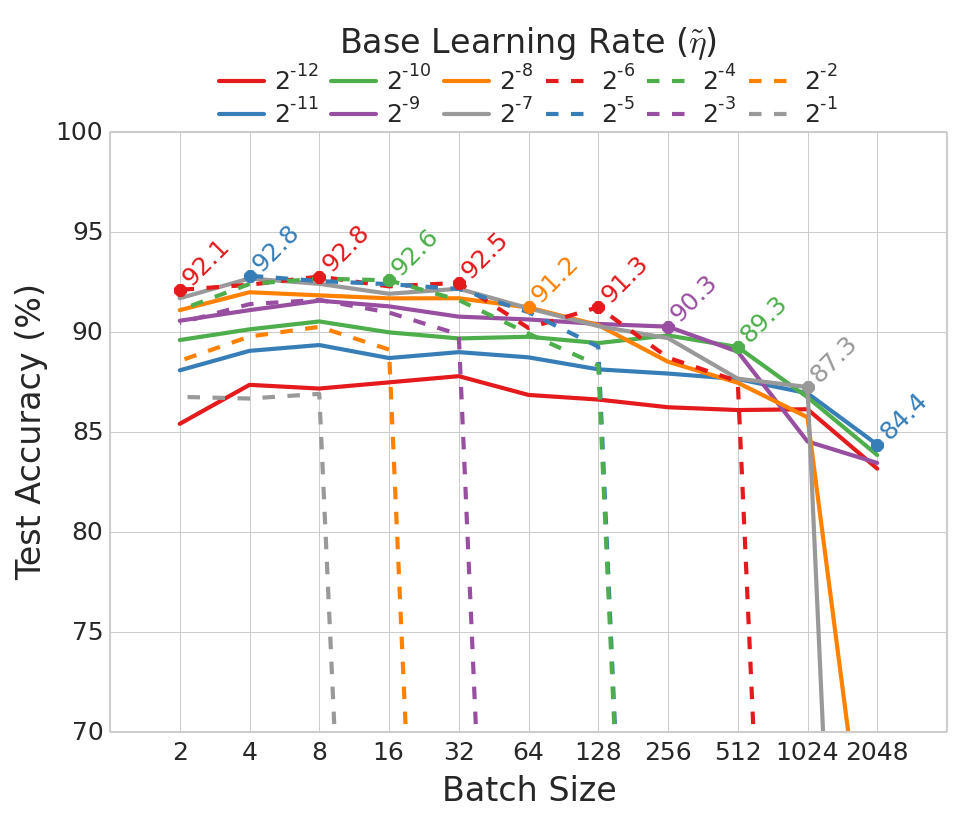}}
\subfigure{\includegraphics[width=0.45\linewidth]{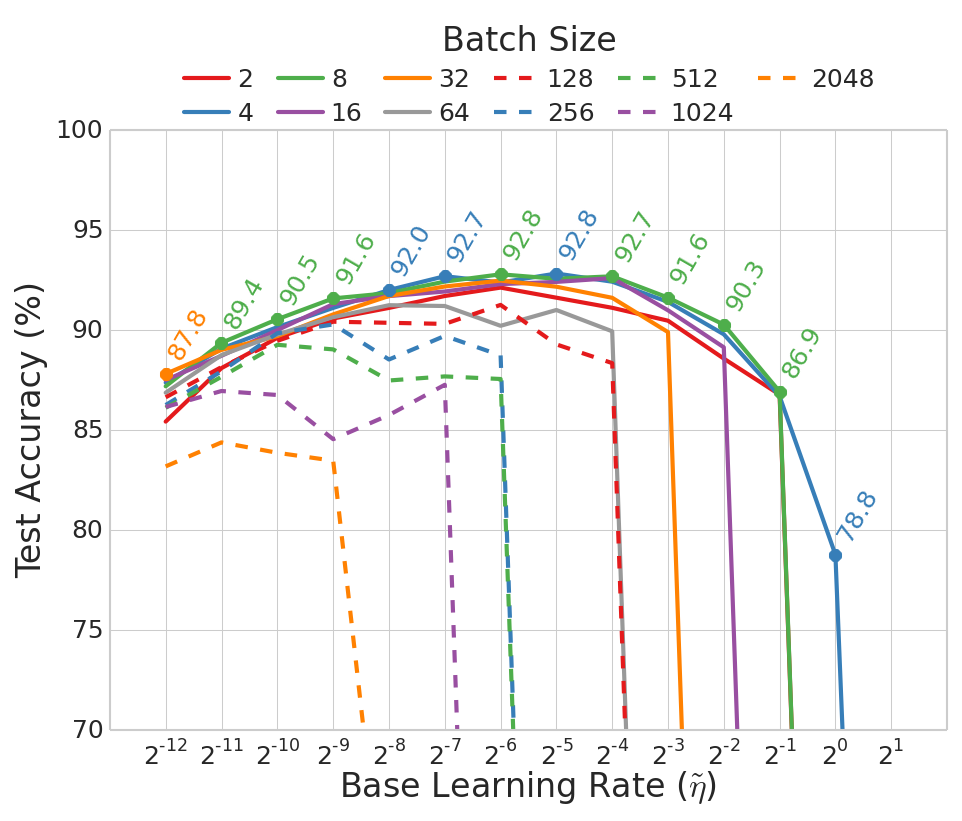}}
\end{subfigmatrix}
\caption{Test performance of ResNet-32 model with BN, for increased training length in number of epochs. CIFAR-10 dataset with data augmentation.}
\label{fig:ResNet-32_CIFAR-10_Aug_BN_2}
\end{figure}

Figures~\ref{fig:ResNet-32_CIFAR-10_Aug_BN_05} and~\ref{fig:ResNet-32_CIFAR-10_Aug_BN_2} show the effect of the training length in number of epochs on the results of Section~\ref{sec:Performance_BN}. The CIFAR-10 ResNet-32 experiments with data augmentation have been repeated for half the number of epochs (41 epochs) and for twice the number of epochs (164 epochs). In both cases, the learning rate reductions have been implemented at 50\% and at 75\% of the total number of epochs. 

The collected results show a consistent performance improvement for increasing training length. For lower base learning rates, the improvements are more pronounced, both for very small batch sizes and for large batch sizes. However, for the base learning rates corresponding to the best test accuracy, the benefits of longer training are important mainly for large batch sizes, in line with~\cite{Hoffer17}.
Even increasing the training length, and therefore the total computational cost, the performance of large batch training remains inferior compared to the small batch performance.

\vspace{0.5cm}

\end{document}